\documentclass[10pt,twocolumn]{article}
\usepackage[letterpaper,
    textheight=8.875in,
    textwidth=6.875in,
    top=1in,
    headheight=0in,
    headsep=0in,
    columnsep=0.3125in]{geometry}
\usepackage{times}
\usepackage{graphicx}
\usepackage{amsmath,amssymb,amsfonts}
\usepackage{booktabs}
\usepackage{xcolor}
\newcommand{\etal}{\textit{et al.}}
\newcommand{\eg}{\textit{e.g.}}
\usepackage[ruled,vlined]{algorithm2e}

\SetCommentSty{mycommfont}
\newcommand{\liying}[1]{{\color{black}{#1}}}

\newcommand{\best}[1]{\textbf{#1}}
\newcommand{\second}[1]{\underline{#1}}
\usepackage{colortbl}
\usepackage{pifont}
\usepackage{arydshln}
\usepackage{multirow,multicol,xspace}
\usepackage{adjustbox}
\usepackage{makecell}
\usepackage{bm}
\makeatletter
\def\adl@drawiv#1#2#3{%
        \hskip.5\tabcolsep
        \xleaders#3{#2.5\@tempdimb #1{1}#2.5\@tempdimb}%
                #2\z@ plus1fil minus1fil\relax
        \hskip.5\tabcolsep}
\newcommand{\cdashlinelr}[1]{%
  \noalign{\vskip\aboverulesep
           \global\let\@dashdrawstore\adl@draw
           \global\let\adl@draw\adl@drawiv}
  \cdashline{#1}
  \noalign{\global\let\adl@draw\@dashdrawstore
           \vskip\belowrulesep}}
\makeatother

\definecolor{lightcyan}{rgb}{0.92, 1.0, 1.0}
\definecolor{lightpurple}{rgb}{0.934, 0.882, 0.937}
\definecolor{lightgreen}{rgb}{0.923, 0.969, 0.912}
\usepackage{hyperref}
\hypersetup{
    colorlinks=true,
    linkcolor=blue!60!black,
    citecolor=green!50!black,
    urlcolor=blue!60!black,
}
\usepackage[numbers,sort&compress]{natbib}
\usepackage[capitalize]{cleveref}
\usepackage{titlesec}
\font\elvbf = ptmb scaled 1100  
\font\tenbf = ptmb scaled 1000  
\usepackage{microtype}
\titleformat{\section}
  {\Large\bfseries}
  {\thesection}
  {0.5em}
  {}
  
\titleformat{\subsection}
  {\elvbf}
  {\thesubsection}
  {0.5em}
  {}
\titleformat{\subsubsection}
  {\tenbf}
  {\thesubsubsection}
  {0.5em}
  {}
\titlespacing*{\section}{0pt}{10pt plus 2pt minus 2pt}{7pt}
\titlespacing*{\subsection}{0pt}{8pt plus 2pt minus 2pt}{5pt}
\titlespacing*{\subsubsection}{0pt}{6pt plus 2pt minus 2pt}{3pt}
\def\xabstract{abstract}
\long\def\abstract#1\end#2{\def\two{#2}\ifx\two\xabstract
    \long\gdef\theabstract{\ignorespaces#1}
    \def\go{\end{abstract}}\else
\typeout{^^J^^J PLEASE DO NOT USE ANY \string\begin\space \string\end^^J
    COMMANDS WITHIN ABSTRACT^^J^^J}#1\end{#2}
\gdef\theabstract{\vskip12pt BADLY FORMED ABSTRACT: PLEASE DO
NOT USE {\tt\string\begin...\string\end} COMMANDS WITHIN
THE ABSTRACT\vskip12pt}\let\go\relax\fi
\go}

\makeatletter
\renewcommand{\maketitle}{%
  \twocolumn[
    \begin{@twocolumnfalse}
      \null
      \vskip .375in
      \begin{center}
        {\Large\bfseries \@title \par}
        \vspace*{24pt}
        {\large
          \lineskip .5em
          \begin{tabular}[t]{c}
            \@author
          \end{tabular}
          \par
        }
        \vskip .5em
        \vspace*{12pt}
      \end{center}
      \centerline{\large\bfseries Abstract}
      \vspace*{12pt}
      {\itshape\theabstract}
      \vspace*{12pt}
    \end{@twocolumnfalse}
  ]
  \thispagestyle{empty}
}
\makeatother

\title{2-Shots in the Dark: Low-Light Denoising with Minimal Data Acquisition}
\author{
    Liying Lu \hspace{20pt} Rapha\"el Achddou \hspace{20pt} Sabine S\"usstrunk \\  
    IVRL, EPFL \\  
    {\tt\small liying.lu@epfl.ch \quad raphael.achddou@esiee.fr \quad sabine.susstrunk@epfl.ch}
}
\date{}

\begin{document}

\begin{abstract}
Raw images taken in low-light conditions are very noisy due to low photon count and sensor noise. Learning-based denoisers have the potential to reconstruct high-quality images. For training, however, these denoisers require large paired datasets of clean and noisy images, which are difficult to collect. Noise synthesis is an alternative to large-scale data acquisition: given a clean image, we can synthesize a realistic noisy counterpart. In this work, we propose a general and practical noise synthesis method that requires only \textbf{one single noisy image and one single dark frame} per ISO setting. We represent signal-dependent noise with a Poisson distribution and introduce a Fourier-domain spectral sampling algorithm to accurately model signal-independent noise. The latter generates diverse noise realizations that maintain the spatial and statistical properties of real sensor noise. As opposed to competing approaches, our method neither relies on simplified parametric models nor on large sets of clean-noisy image pairs.  Our synthesis method is not only accurate and practical, it also leads to state-of-the-art performances on multiple low-light denoising benchmarks.

\end{abstract}

\maketitle

\section{Introduction}
\label{sec:intro}

Capturing images in low-light conditions with consumer cameras often results in extremely noisy photographs. 
To restore clean images from noisy ones, numerous denoising techniques have been developed, including deep learning-based methods~\cite{zhang2017beyond,zamir2022restormer,liang2021swinir,mao2016image,chen2018learning,brooks2019unprocessing}. For such models, the quality of training data is crucial, as it largely determines model effectiveness. Consequently, acquiring appropriate training data is a central challenge.

A common approach is to collect large datasets consisting of clean-noisy image pairs~\cite{chen2018learning,abdelhamed2018high,anaya2018renoir,flepp2024real}. Each pair is obtained by capturing the same scene with different camera settings using a tripod and remote control to obtain well-aligned pairs. This cumbersome process is repeated for multiple scenes.

Noise synthesis has therefore become an active research area, as it allows bypassing the acquisition of such large datasets. However, this task is harder than it appears. While the assumption of Additive White Gaussian Noise (AWGN) has been used in the vast majority of the image denoising literature~\cite{buades2005non,gu2014weighted,zhang2017beyond,zhang2018ffdnet,zhang2021plug}, this model fails to replicate the complex nature of real-world camera noise, resulting in poor performance for denoisers trained under this assumption.

\begin{figure}[t]
\centering
\includegraphics[width=1.0\linewidth]{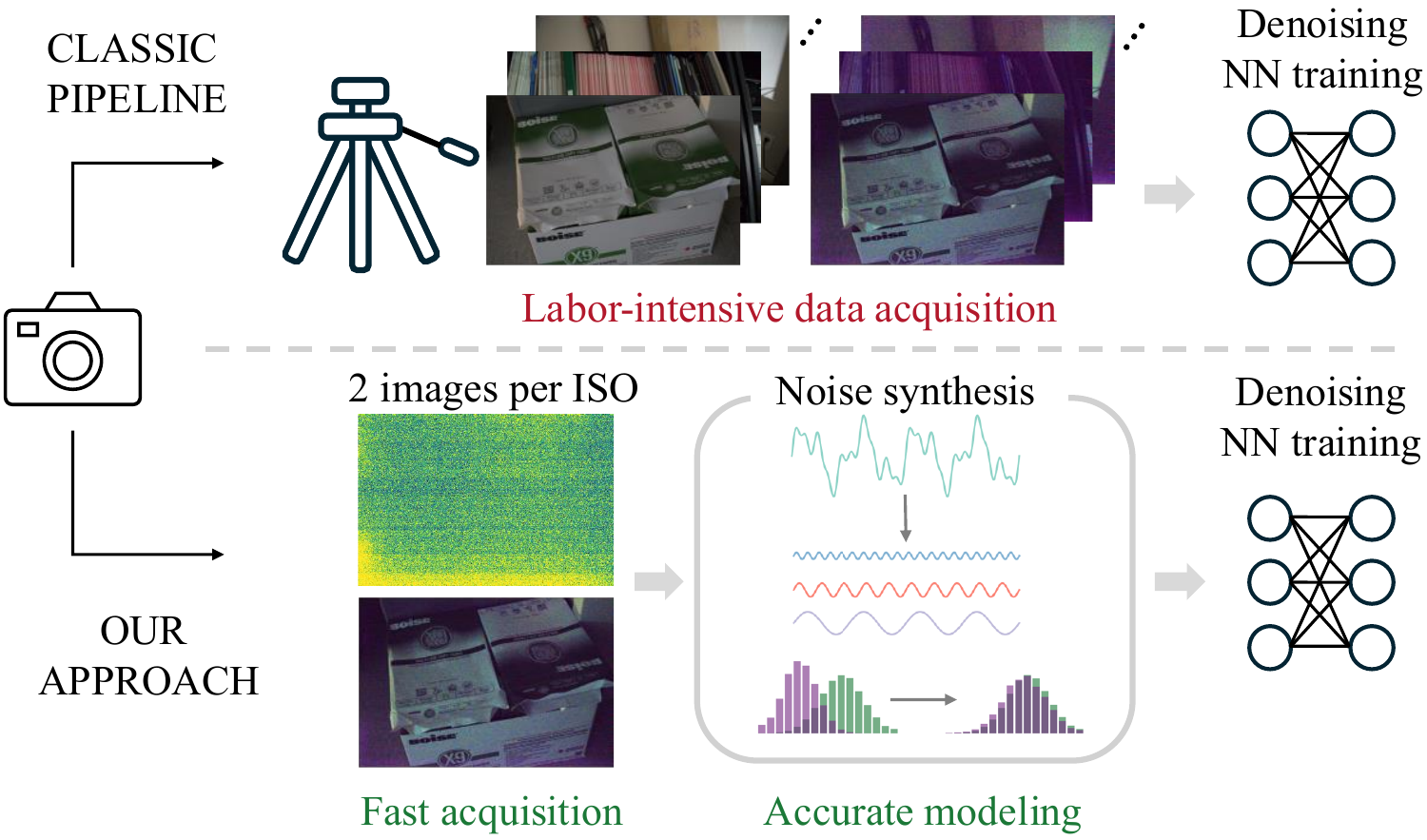}
\caption{Different from classic denoising pipelines that require large amounts of paired data, we propose a precise sensor noise synthesis method that requires only a single noisy image and a single dark frame. From these two inputs, our algorithm accurately reproduces the sensor noise distribution without the need for complex parameter calibration.
}
\label{fig:teaser}
\end{figure}

In practice, sensor noise is the sum of multiple noise sources, which can be divided into signal-dependent and signal-independent components. While signal-dependent noise mainly consists of photon shot noise, signal-independent noise is far more complex and difficult to model. It includes dark current noise~\cite{konnik2014high}, thermal noise~\cite{gow2007comprehensive}, reset noise~\cite{costantini2004virtual,konnik2014high}, and banding pattern noise~\cite{gow2007comprehensive,costantini2004virtual,wei2021physics}.

Prior methods~\cite{costantini2004virtual,wei2021physics} modeled these noise sources with simple parametric models. These methods face two critical limitations: their lack of expressivity and the need for careful parameter calibration. Another direction consists of learning the noise distribution using neural networks, including GANs~\cite{monakhova2022dancing,zhang2023towards}, normalizing flows~\cite{abdelhamed2019noise}, and diffusion models~\cite{lu2025dark}. These approaches have demonstrated significant improvements but still rely on large amounts of paired clean-noisy data for learning the noise distribution.

In this paper, we propose an accurate sensor noise synthesis method that requires only \textit{a single noisy image and a single dark frame}, with the latter captured in a lightless environment. From these two inputs, our algorithm reproduces the sensor noise distribution without the need for complex parameter tuning, significantly reducing the time and effort required for data acquisition and calibration.

Similar to prior works~\cite{wei2021physics,zhang2021rethinking,feng2023learnability}, we model signal-dependent noise with a Poisson distribution, whose parameters can be estimated from the noisy image via straightforward linear regression. Our main contribution lies in the modeling of signal-independent noise. We introduce a Fourier-domain spectral sampling method that is both practical and general. Given a single dark frame, we compute its Fourier transform and apply phase randomization while preserving the magnitude spectrum. This process generates novel noise realizations that maintain the spectral characteristics of the original sensor noise. Meanwhile, histogram matching is employed to ensure that the marginal distribution of the synthetic noise aligns with the real noise.

Our method provides three key advantages for signal-independent noise modeling:
(1) It is a unified framework for modeling all signal-independent noise components; no labor-intensive calibration or acquisition is needed.
(2) It is a general solution applicable to different types of sensors.
(3) Generated noise preserves both the spectral characteristics and marginal distribution of real sensor noise, ensuring a realistic synthesis.

Overall, the proposed method provides an accurate and practical solution with minimal data acquisition for sensor noise synthesis, enabling the generation of unlimited clean–noisy pairs for training image denoising models. Our experiments show that denoising networks trained with our synthetic data are on par with the state-of-the-art models.

\section{Related Work}
\label{sec:related_work}

\subsection{Physics-Based Noise Modeling}\label{sec:physics}
A more realistic alternative to the AWGN approximation of noise used in the vast majority of the denoising literature is the Poisson–Gaussian (PG) model~\cite{foi2008practical,foi2009clipped,brooks2019unprocessing,bahler2022pogain}. This model decomposes noise into two components: a signal-dependent Poisson noise arising from the stochastic nature of photon capture, and a signal-independent Gaussian noise approximating sensor readout noise.

While this model captures fundamental aspects of sensor noise, it does not fully represent the diversity of noise sources encountered in practice. More sophisticated models~\cite{wach2004noise,costantini2004virtual,wei2021physics,zhang2017improved,foi2009clipped} were therefore proposed to incorporate additional noise sources. For instance, the ELD ~\cite{wei2021physics} paper models the read noise with a Tukey-Lambda distribution~\cite{joiner1971some} to better account for its long-tailed profile and color-biases.
Other noise components, such as banding noise, are also considered.
ELD estimates noise parameters using flat-field frames and  dark frames. 
\textit{Flat-field frames} are photographs of a white sheet placed against a uniformly illuminated wall, whereas \textit{dark frames} are captured in a dark room with the camera lens capped.
Despite the more precise noise modeling, the approach remains oversimplified and requires labor-intensive multi-parameter calibration.
The SFRN paper~\cite{zhang2021rethinking} proposes to synthesize noise by directly sampling real dark frames from the sensor.
It models signal-dependent noise using a Poisson distribution, and samples patches from a dark-frame database containing 10 dark frames per ISO setting to mimic signal independent noise.
Our approach shares some similarities with theirs, as we also treat signal-dependent and signal-independent noise separately, and the latter is sampled from a real dark frame exemplar. 
However, there are key differences. Since they crop patches from a small set of dark frames for modeling signal dependent noise, the diversity of the noise samples is somewhat limited. 
In contrast, our spectral sampling method enables the generation of a much more diverse set of noise samples, starting from a single exemplar. 

PMN~\cite{feng2023learnability} proposes applying shot noise augmentation to real clean-noisy pairs. 
It introduces a \textit{dark shading correction} scheme, where hundreds of dark frames are averaged per ISO and linear regression is applied to estimate a combination of black level error (BLE)~\cite{nakamura2017image} and fixed-pattern noise (FPN)~\cite{cain2001projection,holst2007cmos}. 
Subtracting this estimated dark shading from noisy images simplifies subsequent denoising.

\begin{table}[t]
  \caption{\liying{Summary of typical learning-based and non-learning-based noise synthesis methods, along with the number of real image pairs and dark frames used for data synthesis on the SID dataset~\cite{chen2018learning}.}}
  \centering
  \resizebox{1.0\columnwidth}{!}{
  \begin{tabular}{cccc}
    \toprule
    \textbf{Method} & \textbf{Category} & \textbf{\# of real pairs} & \textbf{\makecell[c]{\# of dark frames \\ (per ISO)}} \\ 
    \midrule
    LRD~\cite{zhang2023towards} & Learning & 1865 & 400\footnotemark \\ 
    NoiseDiff~\cite{lu2025dark} & Learning & 1865 & 400\footnotemark  \\ 
    PMN~\cite{feng2023learnability} & Non-learning & 1865 & 400  \\ 
    
    ELD~\cite{wei2021physics} & Non-learning & 0 & Several\footnotemark  \\ 
    SFRN~\cite{zhang2021rethinking} & Non-learning & 0 & 10  \\ 
    Ours & Non-learning & 0 & 1  \\ 
    Poisson-Gaussian & Non-learning & 0 & 0  \\ 
    \bottomrule 
  \end{tabular}}
  \label{tab:method_summary}
\end{table}

\footnotetext[1]{LRD uses the dark shadings from PMN.}
\footnotetext[2]{NoiseDiff uses the dark shadings from PMN.}
\footnotetext[3]{Not mentioned in the ELD paper, but the number is more than one.}

\subsection{Deep Learning-Based Noise Modeling}\label{sec:learning}
Recent studies have explored modeling noise distributions through data-driven approaches based on deep neural networks. Noise Flow~\cite{abdelhamed2019noise} trains a normalizing flow~\cite{dinh2016density,kingma2018glow,kobyzev2020normalizing} to model noise in the RAW domain, and Kousha \etal~\cite{kousha2022modeling} extend flow-based modeling to the sRGB space. LLD~\cite{cao2023physics} proposes to learn different noise components by different normalizing flow layers.
LRD~\cite{zhang2023towards} combines a Poisson model for signal-dependent noise with a GAN for learning signal-independent components. Monakhova~\etal~\cite{monakhova2022dancing} propose a noise generator that learns the parameters of a physics-based model, eliminating manual calibration.
NoiseDiff~\cite{lu2025dark} introduces a diffusion model to learn the sensor noise distribution and achieves superior results. Despite significant progress, these learning-based methods still face several challenges. Normalizing Flow models lack expressiveness for modeling complex distributions because of architectural constraints. GANs are prone to training instability~\cite{gulrajani2017improved,mescheder2018training}. Diffusion models~\cite{ramesh2022hierarchical,rombach2022high} provide more accurate noise modeling but suffer from a relatively slow generation speed. Moreover, all these models still require a certain amount of real paired data for training.
In contrast, our method does not rely on any paired data. It requires only a single noisy image and one dark frame, yet it can generate infinitely diverse samples with high efficiency. \liying{We summarize the amount of data used by different synthesis methods for noise generation in Table~\ref{tab:method_summary}.}

\section{Method} 

\subsection{Spectral Sampling for Dark Frame Synthesis}

We aim to model the complex signal-independent noise, which includes multiple components such as fixed-pattern noise (FPN), dark current shot noise, read noise, and banding pattern noise. Parametric models require laborious sensor-specific calibration, yet still struggle to capture the full complexity of real noise.

We reinterpret this problem from a \textit{texture synthesis} perspective, drawing inspiration from the random phase noise (RPN) algorithm~\cite{galerne2010random,lewis1984texture,van1991spot}. 
The RPN algorithm synthesizes textures by preserving the Fourier magnitude spectrum while randomizing the phase. For stationary stochastic textures, the magnitude encodes spatial correlations and frequency content, while the phase primarily determines spatial localization. 

Interestingly, signal-independent noise in dark frames exhibits approximately stationary properties after removal of the fixed-pattern noise. 
This property makes Fourier-domain synthesis particularly suitable for modeling sensor noise: by preserving the magnitude spectrum from a reference dark frame and randomizing its phase, we can generate new noise realizations that retain the frequency characteristics without requiring explicit parametric modeling.

\begin{figure*}[t]
  \centering
  \includegraphics[width=1.0\linewidth]{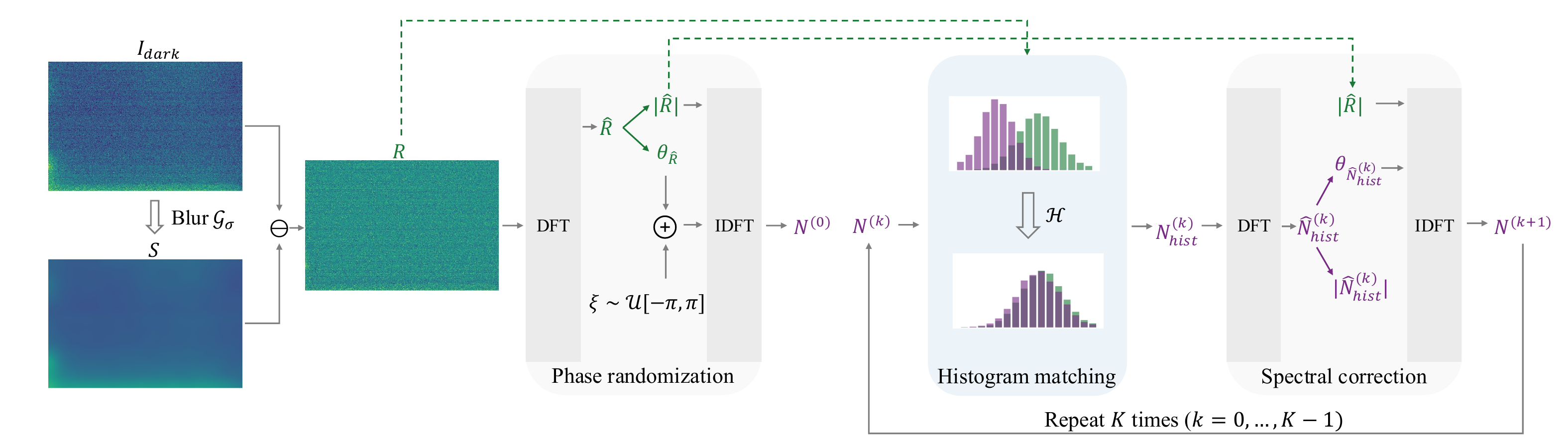}
  \caption{Overview of the spectral sampling algorithm. A Gaussian blur is applied to the real dark frame $I_{\text{dark}}$ to estimate the fixed-pattern component $S$, which is then subtracted to obtain the stochastic residual $R$. In the Fourier domain, we retain the magnitude $|\widehat{R}|$ and perform phase randomization using a uniform random phase $\xi$ to obtain a new noise realization $N^{(0)}$. Subsequently, $K$ iterations of histogram matching and spectral correction are applied to preserve both the marginal distribution and the spectral characteristics of the noise.
    }
  \label{fig:framework}
\end{figure*}

Furthermore, we incorporate iterative histogram matching to ensure that the synthesized noise matches both the spectral characteristics and the full marginal distribution of the reference noise, producing noise realizations that closely reflect real sensor properties. The entire pipeline is illustrated in Fig.~\ref{fig:framework}.

\subsubsection{Fixed-Pattern Removal}
\label{sec:fix_pattern_removal}

Dark frames contain both \textit{structured} low-frequency patterns (e.g., bias shading, fixed-pattern noise) and \textit{stochastic} signal-independent noise. Direct spectral analysis of the dark frame would be dominated by these low-frequency components, biasing the estimated spectral statistics.  
To isolate the stochastic component, we estimate the fixed-pattern structure by applying a large-kernel Gaussian blur to the reference dark frame $I_{\text{dark}}$:
\begin{equation}
    S = \mathcal{G}_{\sigma} * I_{\text{dark}}, \label{eq:gaussian_blur}
\end{equation}
where $\mathcal{G}_{\sigma}$ is a Gaussian kernel of standard deviation $\sigma$. This filtering isolates spatially smooth patterns and removes high-frequency variations.  
Subtracting this estimate yields the residual:
\begin{equation}
    R' = I_{\text{dark}} - S,
\end{equation}
which primarily contains the stochastic noise components. To focus on the noise fluctuations and simplify spectral analysis, we further remove the channel-wise mean:
\begin{align}
    \mu^{R} &= \mathbb{E}[R'], \label{eq:compute_mean}
    R = R' - \mu^{R},
\end{align}
where $\mathbb{E}[\cdot]$ is the channel-wise spatial averaging operator.

\subsubsection{Spectral Prior Estimation}
We then compute the 2D discrete Fourier transform (DFT) of the zero-mean noise residual $R$:
\begin{equation}
    \widehat{R} = \mathcal{F}\{ R\},
\end{equation}
where $\mathcal{F}\{\cdot\}$ denotes the DFT applied independently to each color channel. The magnitude spectrum 
$|\widehat{R}|$
encodes the characteristic frequency content and spatial correlation patterns of the sensor noise, serving as a 
\emph{spectral prior} for subsequent synthesis.

\subsubsection{Phase Randomization}

To synthesize new noise realizations that preserve the magnitude spectrum of real signal-independent noise, we employ a phase randomization algorithm inspired by RPN. Starting from the magnitude spectrum$|\widehat{R}|$ obtained from the reference residual noise, we introduce randomness by incrementing the phase with a random offset $\xi \in \mathbb{R}^{C \times H \times W}$.

$\xi$ is obtained by first creating a single channel map $\xi^0$ sampled from a uniform distribution over $[-\pi,\pi]$, and replicating it across all color channels. Given $\xi$, we construct a new randomized spectrum $\widehat{N}$:
\begin{align}
     \xi^0 &\sim \mathcal{U} \big( [-\pi, \pi]^{(H,W)} \big),\label{eq:random_phase} \\
     \xi &= \text{replicate}(\xi^0, C),\label{eq:replicate} \\
    \widehat{N} &= |\widehat{R}| \odot \exp\Big(i \big(\theta_{\widehat{R}} +  \xi\big)\Big), \label{eq:rpn}
\end{align}
where $\theta_{\widehat{R}}$ denotes the phase of $\widehat{R}$, and $\odot$ represents element-wise multiplication. This operation preserves the magnitude spectrum $|\widehat{R}|$ while randomizing the phase to generate diverse noise instances.

A subtle but critical design choice is to replicate the same values across all channels of the phase offset $\xi$. 
In real sensor data, certain noise components such as banding patterns exhibit strong \textit{inter-channel correlation}. 
Ensuring all channels share an identical random phase preserves these correlations. 
In contrast, applying independent phase offsets per channel would break them, producing unrealistic noise and leading to residual artifacts in images processed by denoisers trained on such synthetic data, as discussed in \cref{sec:discussions}.

\subsubsection{Inverse Transform and Initial Reconstruction}
The synthesized noise realization is obtained by applying the inverse Fourier transform with proper normalization:
\begin{equation}
    N^{(0)} = \frac{1}{\sqrt{HW}} \mathcal{F}^{-1}\{\widehat{N}\},
\end{equation}
where $H$ and $W$ are the height and width of $\widehat{N}$.

Although phase randomization preserves channel correlations and the magnitude spectrum, it does not guarantee the preservation of the reference noise histogram. Real sensor noise often exhibits non-Gaussian characteristics, including asymmetry and heavy tails~\cite{wei2021physics}, which the magnitude spectrum alone cannot capture. Ignoring these properties can lead to synthesized noise that appears visually plausible but does not accurately reproduce important noise statistics, such as mean, variance, skewness, and kurtosis, all of which are essential for training robust denoising models, as we demonstrate in \cref{sec:discussions}.

\subsubsection{Iterative Histogram and Spectral Refinement}
To address this limitation, we introduce an iterative refinement procedure that alternates between histogram matching and spectral constraint enforcement. This approach ensures that the synthesized noise simultaneously satisfies both the spectral characteristics and the complete pixel-wise marginal distribution of the reference noise.

The refinement proceeds as follows. At iteration $k$, we first apply histogram matching to align the marginal distribution of $N^{(k)}$ with that of the reference residual $R$:
\begin{equation}
    N'^{(k)}_{\text{hist}} = \mathcal{H}(N^{(k)}, R),
\end{equation}
where $\mathcal{H}(\cdot, \cdot)$ denotes the histogram matching operator. 

However, histogram matching may distort the frequency content and channel correlations. 
To restore these properties, we re-impose the spectral constraint:
\begin{align}
    N^{(k)}_{\text{hist}} &= N'^{(k)}_{\text{hist}} - \mu^{(k)}_{\text{hist}}, \\
    \widehat{N}^{(k)}_{\text{hist}} &= \mathcal{F}\{N^{(k)}_{\text{hist}}\}, \\
    \widehat{N}^{(k)}_{\text{corrected}} &= |\widehat{R}| \odot \exp\big(i \theta_{\widehat{N}^{(k)}_{\text{hist}}}\big), \\
    N^{(k+1)} &= \frac{1}{\sqrt{HW}} \mathcal{F}^{-1}\{\widehat{N}^{(k)}_{\text{corrected}}\} + \mu^{(k)}_{\text{hist}},
\end{align}
where $\mu^{(k)}_{\text{hist}}$ is computed in a similar way to $\mu^R$ in Eq.~\ref{eq:compute_mean}.
By preserving the phase from the histogram-matched sample while enforcing the reference magnitude spectrum, we maintain spatial randomness while restoring the correct spectral characteristics. To summarize, this iterative process alternates between two complementary constraints:
\begin{itemize}
    \item \textbf{Histogram matching} ensures that the pixel-wise marginal distribution matches the reference.
    \item \textbf{Spectral correction} ensures that the frequency characteristics and spatial patterns match the reference.
\end{itemize}

\subsubsection{Final Reconstruction}
After $K$ iterations, we reintroduce the deterministic low-frequency structure and channel means by adding back the fixed-pattern estimate $S$ and the channel-wise mean $\mu^{R}$:
\begin{equation}
    \tilde{I}_{\text{dark}} = N^{(K)} + S + \mu^{R}.
\end{equation}
This step ensures that the synthesized dark frames retain both the structured sensor biases and the realistic stochastic noise characteristics observed in real data.

We summarize the complete procedure in Algorithm 1 in the appendix.

\subsection{Photon Noise Parameter Estimation}
\label{sec:shot_noise_est}

Photon noise can be approximated by the Poisson distribution because of the quantum nature of photons. To estimate the gain parameter $g$, we follow a similar approach as \cite{healey2002radiometric,wei2021physics,feng2023learnability}. Generally, the noisy observation can be modeled as: $y = g\mathcal{P}(x) + n_{\text{other}}$,
where $x$ is the latent clean signal, $\mathcal{P}$ denotes the Poisson distribution and $n_{\text{other}}$ denotes signal-independent noise components. This would lead to a linear relationship between the variance of the noisy observation $y$ and the amplified clean signal $gx$:
\begin{equation}
\text{Var}(y)=g (gx) + \text{Var}(n_{\text{other}}) \label{eq:photon_var},
\end{equation}
and a linear regression can be applied to estimate $g$.
If only a single noisy image is available, a rough estimate of $g$ can be obtained by first applying a Gaussian blur to approximate a pseudo clean image, and then using paired pixels from the noisy and pseudo clean images to fit the linear relationship. More details can be found in the appendix.

\begin{table*}[t]
    \caption{Denoising performance on the SID and ELD datasets in terms of PSNR / SSIM across various exposure ratios. 
    Best results are in \best{bold}, and second-best results are \second{underlined}. \textit{PG} denotes Poisson-Gaussian. \textit{Real data} denotes the results obtained by using real paired data. \textit{LRD} and \textit{NoiseDiff} are learning-based methods, while others are \liying{non-learning-based} methods. LRD, NoiseDiff and PMN require large numbers of real clean-noisy pairs to synthesize new data, while ours requires only one noisy image and one dark frame per ISO.
    }
  \centering
  \resizebox{1.0\textwidth}{!}{
  \begin{tabular}{ccc >{\columncolor{white}}c >{\columncolor{white}}c >{\columncolor{gray!10}}c >{\columncolor{gray!10}}c >{\columncolor{gray!10}}c >{\columncolor{gray!10}}c >{\columncolor{gray!10}}c }
  
  \toprule
  \textbf{Dataset}  & \textbf{Ratio} & \cellcolor{white} \textit{\textbf{Real data}} & \cellcolor{white} \textbf{LRD}~\cite{zhang2023towards} & \cellcolor{white} \textbf{NoiseDiff~\cite{lu2025dark}} & \cellcolor{white} \textbf{PG} & \cellcolor{white} \textbf{ELD}~\cite{wei2021physics} &  \cellcolor{white} \textbf{SFRN}~\cite{zhang2021rethinking} & \cellcolor{white} \textbf{PMN}~\cite{feng2023learnability} & \cellcolor{white} \textbf{Ours} \\ 
  \midrule
   
    \multirow{3}{*}{\textbf{SID}} 
    & $\times$100 & \textit{42.95 / 0.958} & 43.16 / 0.958 & \best{43.92 / 0.961} & 41.05 / 0.936 & 41.95 / 0.953 & 42.81 / 0.957 & 43.47 / \best{0.961} & \second{43.57} / \best{0.961} \\ 
    & $\times$250 & \textit{40.27 / 0.943} & 40.69 / 0.941 & \best{41.28} / \second{0.946} & 36.63 / 0.885 & 39.44 / 0.931 & 40.18 / 0.934 & 41.04 / \best{0.947} & \second{41.24} / 0.945  \\ 
    & $\times$300 & \textit{37.32 / 0.928} & 37.48 / 0.919 & \best{37.90} / \second{0.929} & 33.34 / 0.811 & 36.36 / 0.911 & 37.09 / 0.918 & \second{37.87} / \best{0.934} & 37.77 / \second{0.929}  \\ 
    
    \cdashlinelr{1-10}
    \multirow{2}{*}{\textbf{ELD}} 
    & $\times$100 & \textit{45.52 / 0.977} & 46.16 / 0.983 & 46.95 / 0.978 & 44.28 / 0.936 & 45.45 / 0.975 & 46.38 / 0.979 & \second{46.99 / 0.984} & \best{47.13 / 0.986}  \\ 
    & $\times$200 & \textit{41.70 / 0.912} & 43.91 / 0.968 & \best{45.11 / 0.971} & 41.16 / 0.885 & 43.43 / 0.954 & 44.38 / 0.965 & 44.85 / \second{0.969} & \second{44.89 / 0.969} \\
   \bottomrule 
  \end{tabular}}
\label{tab:compare_sid}
\end{table*}

\section{Experiments}

\subsection{Compared Methods}

To assess the quality of the synthetic data, we compare the performance of denoising networks trained with data from different synthesis methods.
We evaluate two recent learning-based methods (LRD~\cite{zhang2023towards} and NoiseDiff~\cite{lu2025dark}), and four physics-based methods (the Poisson-Gaussian model, ELD~\cite{wei2021physics}, SFRN~\cite{zhang2021rethinking} and PMN~\cite{feng2023learnability})\footnote{For LRD, NoiseDiff, ELD and PMN, we use pre-trained models available on the authors’ official websites. For SFRN, we use the reproduced code from the PMN repository whose results have been confirmed by the SFRN authors.}.
We also compare with a denoising network trained on real data pairs from the SID Sony training set. 
Among these methods, LRD, NoiseDiff and PMN still require large numbers of real clean-noisy pairs to synthesize new data. SFRN, ELD and our method synthesize data without the help of paired data.

\subsection{Implementation Details}

For noise synthesis, we use images from the SID Sony training set~\cite{chen2018learning}, captured with a Sony A7S2 camera. 
We additionally use the dark-frame dataset from LLD~\cite{cao2023physics}, which contains 400 dark frames per ISO level, captured with the same camera model (though different physical devices).
For each ISO level, we use one dark frame from the LLD dataset and one noisy image from the SID training set to synthesize dark frames.
From these two images, we generate 400 synthetic dark frames for that ISO.

During denoising training, we take clean images from the SID training set and synthesize noisy images by applying Poisson noise and adding a randomly selected synthetic dark frame corresponding to the same ISO.

Among the compared methods, LRD, NoiseDiff, and PMN perform dark shading correction (DSC), where a precomputed dark shading map is subtracted from the noisy inputs during both denoising training and inference (see Sec.~\ref{sec:physics}). To ensure a fair comparison, we also train and test our denoiser with DSC. However, since we assume access to only a single dark frame per ISO, we estimate the dark shading map by applying Gaussian smoothing to that frame, as described in Sec.~\ref{sec:fix_pattern_removal}.
More implementation details can be found in the appendix.

\subsection{Image Denoising Performance Comparison}

For denoising performance evaluation, we report results on the SID and ELD test sets.
Table~\ref{tab:compare_sid} reports the denoising performance of different methods. Although our approach synthesizes training data using only one noisy image and one dark frame per ISO, the denoising network trained on our synthetic data achieves strong performance across all exposure ratios on the SID and ELD test sets.

Among all methods, \textit{NoiseDiff} achieves the best results on the SID dataset. However, it is a learning-based noise model that requires a large collection of clean–noisy pairs to learn the noise distribution \liying{(as shown in Table~\ref{tab:method_summary})}, and both its diffusion-based training and inference are computationally expensive. In contrast, among physics-based methods, our approach consistently achieves the best performance on the SID and ELD test sets across most exposure levels.

\begin{figure*}[t]
\centering
\includegraphics[width=1\linewidth]{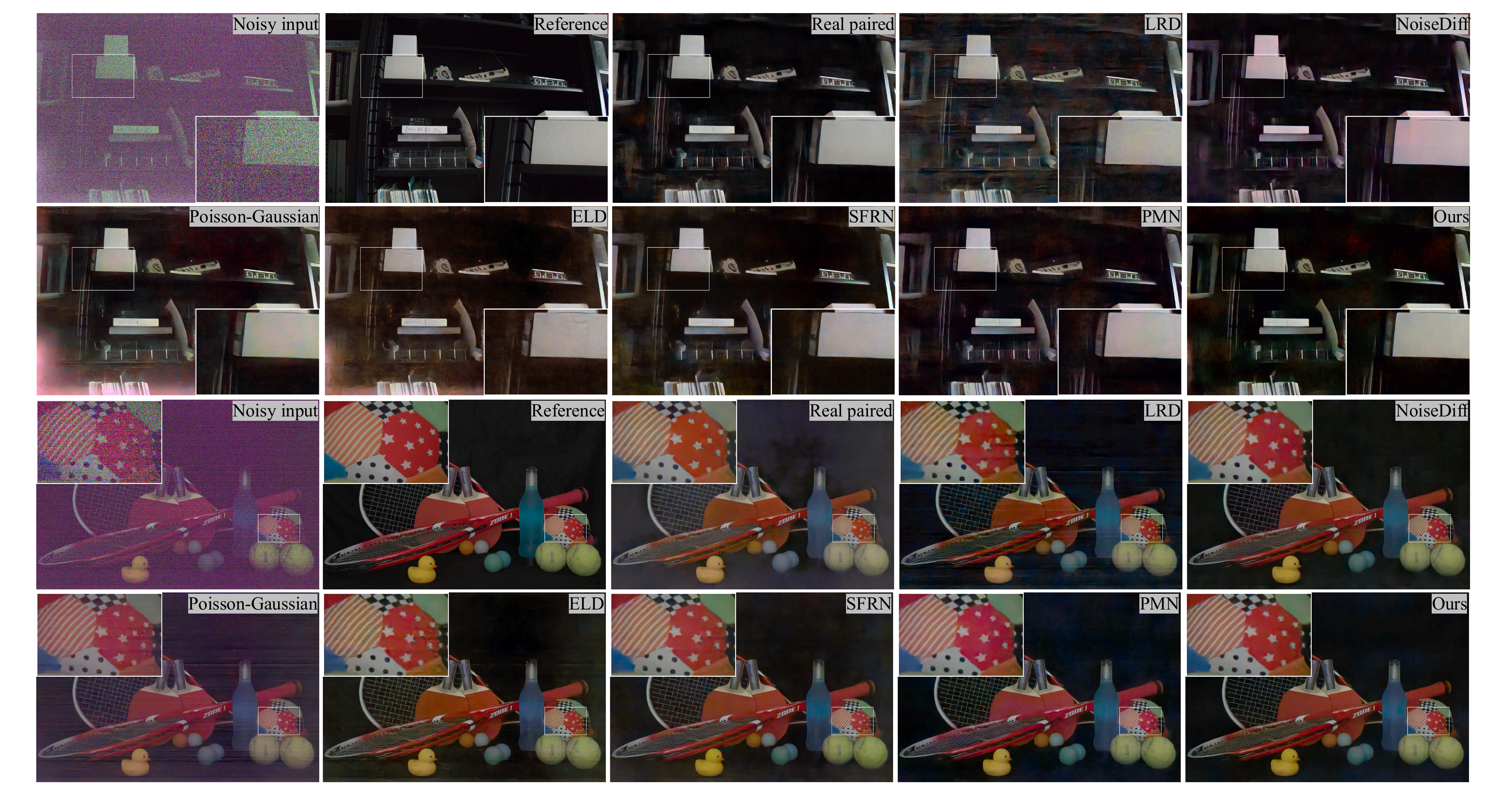}
\caption{Result comparison of denoisers trained on data synthesized using different methods, along with a comparison to a denoiser trained on real pairs from the SID training set. The first example is from the SID test set, the second from the ELD test set. 
For both examples, our method produces cleaner results with fewer artifacts than other methods.
Best viewed zoomed in. More results are provided in the appendix.
}
\label{fig:result_compare_sid_eld}
\vspace{-0.5mm}
\end{figure*}

Figure~\ref{fig:result_compare_sid_eld} presents visual comparisons. In the first example from the SID test set, our method produces a cleaner reconstruction compared to other physics-based methods. In the second example from the ELD test set, our method effectively removes banding noise, whereas PG, ELD, and LRD leave noticeable residual banding. PMN also leaves visible noise in dark regions, and \textit{NoiseDiff} introduces color deviations in the orange stripes.

\subsection{Generalization to Different Sensors}

To evaluate the generality of our approach, we further apply it to a different sensor using the LRID dataset~\cite{feng2023learnability}, which is captured with a Redmi K30 smartphone (IMX686 sensor). We follow the same data synthesis procedure as in the SID experiments, using one noisy image and one dark frame per ISO to synthesize noise.

Table~\ref{tab:compare_lrid} reports the quantitative results on the LRID test set. We primarily compare against state-of-the-art physics-based noise modeling methods. Our method consistently achieves superior performance on the LRID-Indoor test set and remains competitive on the LRID-Outdoor test set. While PMN attains higher performance on the LRID-Outdoor test set, it benefits from using hundreds of dark frames to estimate dark shading, which provides a strong prior and effectively stabilizes denoising training. Moreover, PMN still relies on a large number of real clean–noisy pairs for network training \liying{as shown in Table~\ref{tab:method_summary}.}

In contrast, our approach requires only a single noisy image and a single dark frame per ISO, yet achieves strong generalization across sensors without the need for extensive dark frame collections or clean–noisy datasets. This result suggests that our noise synthesis algorithm is scalable across various image sensors. Visual comparisons on the LRID dataset can be found in the appendix.

\begin{table}[t]
    \caption{Denoising results on the LRID-Indoor (top) and -Outdoor (bottom) datasets in terms of PSNR / SSIM across various exposure ratios. }
  \centering
  \resizebox{1.0\columnwidth}{!}{
  \begin{tabular}{cc >{\columncolor{gray!10}}c >{\columncolor{gray!10}}c >{\columncolor{gray!10}}c >{\columncolor{gray!10}}c >{\columncolor{gray!10}}c }
  
  \toprule
  \textbf{Ratio} & \cellcolor{white} \textit{\textbf{Real data}} & \cellcolor{white} \textbf{PG} & \cellcolor{white} \textbf{ELD}~\cite{wei2021physics} &  \cellcolor{white} \textbf{SFRN}~\cite{zhang2021rethinking} & \cellcolor{white} \textbf{PMN}~\cite{feng2023learnability} & \cellcolor{white} \textbf{Ours} \\ 
  \midrule
   
    $\times$64 & \textit{48.80 / 0.991} & 46.98 / 0.988 & 48.26 / 0.990 & 48.18 / 0.990 & \second{49.32} / \best{0.992} & \best{49.47 / 0.992}  \\ 
    $\times$128 & \textit{47.10 / 0.986} & 45.93 / 0.982 & 46.69 / 0.984 & 46.75 / 0.986 & \second{47.60} / \best{0.987} & \best{47.72 / 0.987} \\ 
    $\times$256 & \textit{44.89 / 0.979} & 44.09 / 0.970 & 44.47 / 0.974 & 44.84 / 0.979 & \second{45.41} / \best{0.981} & \best{45.50} / \second{0.979} \\ 
    $\times$512 & \textit{42.59 / 0.966} & 41.55 / 0.946 & 41.78 / 0.947 & 42.69 / 0.966 & \second{43.14} / \best{0.967} & \best{43.23} / \second{0.966} \\ 
    $\times$1024 & \textit{40.29 / 0.945} & 38.22 / 0.894 & 38.39 / 0.903 & 40.38 / 0.947 & \second{40.67} / \best{0.948} & \best{40.76} / \second{0.940} \\ 
    
    \cdashlinelr{1-7}  
    $\times$64 & \textit{45.85 / 0.988} & 42.51 / 0.980 & 45.09 / 0.984 & 45.18 / 0.985 & \best{46.32 / 0.988} & \second{46.31} / \best{0.988} \\ 
    $\times$128 & \textit{44.52 / 0.982} & 41.78 / 0.972 & 43.63 / 0.974 & 43.83 / 0.977 & \best{44.90 / 0.983} & \second{44.76 / 0.980} \\ 
    $\times$256 & \textit{42.71 / 0.971} & 40.59 / 0.953 & 41.52 / 0.948 & 42.08 / 0.961 & \best{43.01 / 0.970} & \second{42.72 / 0.958} \\ 
   
   \bottomrule 
  \end{tabular}}
\label{tab:compare_lrid}
\end{table}

\begin{table}[t]
    \caption{Ablation study on inter-channel correlation (ICC) and iterative histogram matching (IHM). Results are reported in PSNR~/~SSIM on the SID and ELD test sets.}
    \centering
    \resizebox{0.87\columnwidth}{!}{
    \begin{tabular}{ccccc}
    \toprule
    \textbf{Dataset}  & \textbf{Ratio} & w/o ICC & w/o IHM & Ours \\ 
    \midrule
    \multirow{3}{*}{\textbf{SID}} 
    & $\times$100 & 43.63 / 0.959 & 43.55 / 0.952 & 43.72 / 0.961  \\ 
    & $\times$250 & 40.94 / 0.935 & 40.75 / 0.926 & 41.30 / 0.944  \\ 
    & $\times$300 & 37.51 / 0.917 & 37.39 / 0.911 & 37.86 / 0.929 \\ 
      
    \cdashline{1-5}
    \multirow{2}{*}{\textbf{ELD}} 
    & $\times$100 & 47.18 / 0.986 & 47.06 / 0.984 & 47.14 / 0.986 \\ 
    & $\times$200 & 44.82 / 0.967 & 44.81 / 0.966 & 44.78 / 0.966 \\ 
      
    \bottomrule 
    \end{tabular}}
    \label{tab:ablation}
\end{table}

\subsection{Discussions}
\label{sec:discussions}

\begin{figure*}[t]
\centering
\includegraphics[width=1\linewidth]{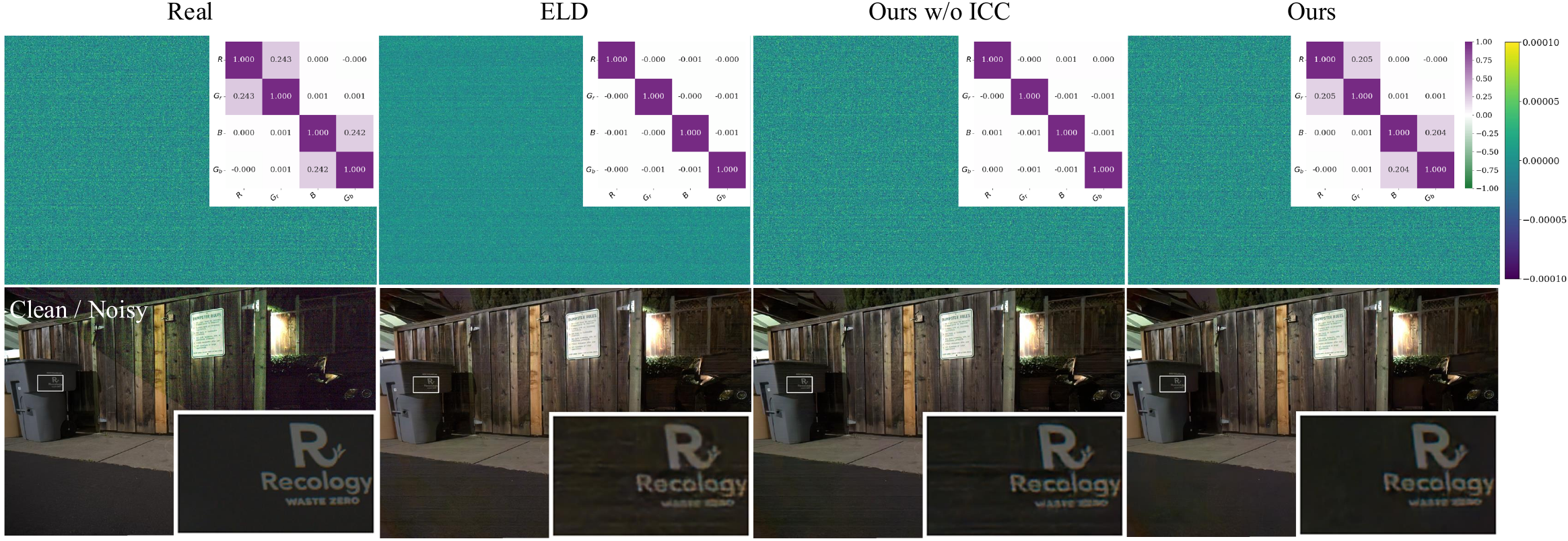}
\caption{Effect of inter-channel correlation (ICC). \textit{First row:} Noise samples (ISO=50) from different methods, with row-wise averaged correlation matrices for Bayer channels (R, Gr, Gb, B). 
\textit{Second row:} A clean-noisy pair and denoising results when denoisers are trained on different with different noise samplers. 
Real sensor noise exhibits cross-channel correlations, but ELD and our method without correlation enforcement produce nearly independent channels (near-diagonal matrices). 
Only our full method replicates the correlations. Consequently, denoisers trained with ELD or without correlation enforcement show residual banding artifacts, while our full method yields clean results.
}
\label{fig:channel_correlation}
\vspace{-2mm}
\end{figure*}

\begin{figure*}[t]
\centering
\includegraphics[width=1\linewidth]{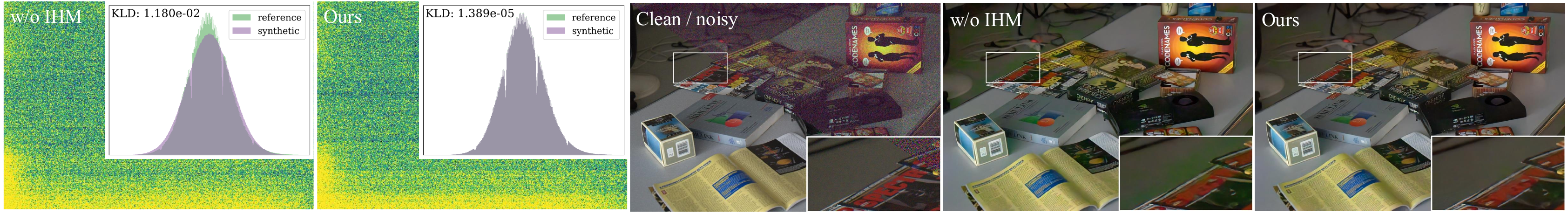}
\caption{Effect of iterative histogram matching (IHM). \textit{Column 1-2}: synthetic noise without and with IHM (with histogram comparison and KLD to real noise). \textit{Column 3}: a real clean-noisy pair. \textit{Column 4-5}: the denoising results when training with data synthesized without and with IHM.
Without IHM, there is a noticeable distribution misalignment between the \textcolor[HTML]{9970ab}{synthetic} and \textcolor[HTML]{5aae61}{real} noise, and the denoising result exhibits color artifacts. Whereas IHM improves noise realism and restoration quality. Best viewed zoomed in.}
\label{fig:hist_iter}
\vspace{-3mm}
\end{figure*}

\begin{figure}[tph]
\centering
\includegraphics[width=1\linewidth]{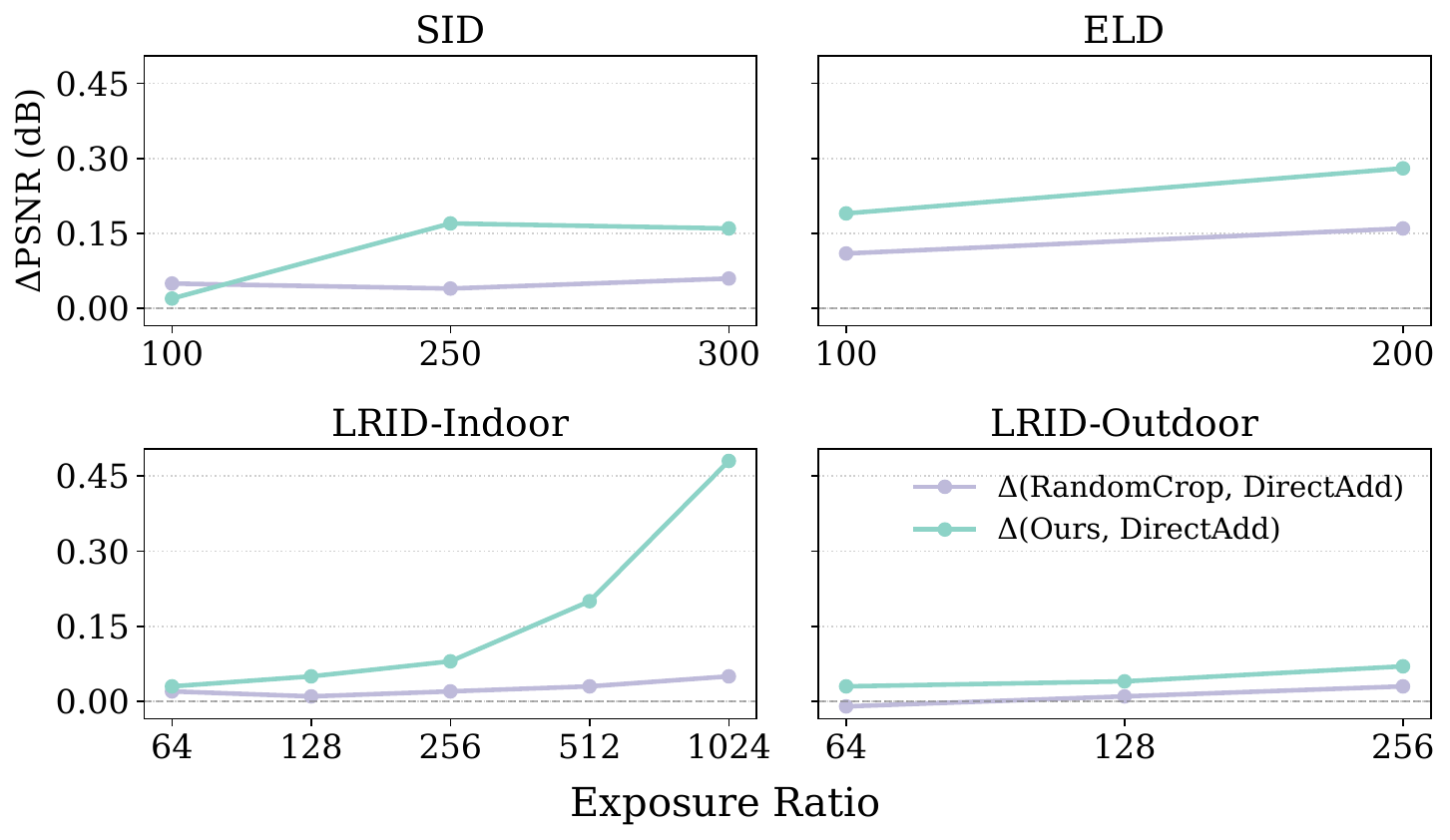}
\caption{PSNR differences relative to the baseline \textit{DirectAdd}: \textit{RandomCrop} vs. \textit{DirectAdd}, and \textit{Ours} vs. \textit{DirectAdd} across exposure ratios on the SID, ELD, and LRID test sets.
}
\label{fig:lineplot_use_one_real}
\vspace{-2mm}
\end{figure}

\noindent \textbf{Importance of inter-channel correlation.}
A critical but often overlooked aspect of sensor noise modeling is inter-channel correlation (ICC). 
Fig.~\ref{fig:channel_correlation} (first row) compares the ICC of real noise and noise synthesized with different methods. We compute Pearson correlation coefficients among the four color channels for each row, then average across all rows. 
Real noise shows clear positive correlations between some channels, while ELD~\cite{wei2021physics} and our method without correlation enforcement produce nearly diagonal matrices, indicating spurious independence. 
Our full method reproduces correlations by using a shared random phase replicated across all channels (Eq.~\ref{eq:random_phase} and \ref{eq:replicate}).

The reproduction of these correlations is crucial for training robust denoisers. The bottom row of Fig.~\ref{fig:channel_correlation} shows that denoisers trained with \textit{ELD} or \textit{Ours w/o ICC} data exhibit visible residual banding artifacts, while our method produces clean results.
Table~\ref{tab:ablation} reports the quantitative results. Removing ICC leads to consistently lower PSNR/SSIM on the SID dataset. The performance on the ELD dataset appears similar; however, as shown in the appendix, ICC remains crucial for suppressing banding noise.

Notably, our approach requires no explicit assumptions about channel or spatial correlations. By preserving the reference magnitude spectrum and controlling phase randomization, it naturally maintains the inherent correlation characteristics of each sensor.


\noindent \textbf{Effect of iterative histogram matching.}
The iterative histogram matching (IHM) module is designed to align the marginal distribution of the synthesized noise with that of real noise. To analyze its effect, we compare synthesized noise with and without IHM. The first two columns of Fig.~\ref{fig:hist_iter} show synthetic dark frames generated with and without IHM. Although their visual appearances are similar, the version without IHM exhibits a noticeable distribution mismatch from real noise.
Columns 4 and 5 of Fig.~\ref{fig:hist_iter} show denoising results produced by networks trained on data synthesized with and without IHM. Without IHM, the reconstructed image exhibit color distortions, indicating that the distribution mismatch in the synthetic noise propagates to the downstream denoiser. With IHM, these artifacts are substantially reduced. We also observe that IHM helps suppress saturated malfunctioning pixels (see appendix), likely because it better preserves the long-tailed profile of the real noise histogram.

Finally, Table~\ref{tab:ablation} summarizes the quantitative results. Removing IHM results in lower metrics, especially on the SID dataset. In all subsequent experiments, we set $K = 10$, as it provides a good balance between synthesis quality and computational cost.

\noindent \textbf{Different ways to leverage a single dark frame.}
When only one real dark frame is available, several strategies can leverage it to generate paired training data, with additional Poisson noise applied to model signal-dependent noise:  
1) \textit{Direct addition:} Add the real dark frame to $M$ clean images, producing $M$ clean–noisy pairs, but all noisy samples share the same noise pattern.  
2) \textit{Patch-based sampling:} Crop random patches from the dark frame and add them to clean patches, as in SFRN~\cite{zhang2021rethinking}, introducing some variation but limited diversity.  
3) \textit{Our method:} Generate multiple dark frame variations, offering much greater diversity.  

We perform an ablation study by training three denoisers on synthetic datasets generated by each method (\textit{DirectAdd}, \textit{RandomCrop}, \textit{Ours}). 
Fig.~\ref{fig:lineplot_use_one_real} illustrates the denoising performance differences relative to the baseline \textit{DirectAdd}. The \textcolor[HTML]{bebada}{purple lines} show the $\Delta\text{PSNR}$ of \textit{RandomCrop} minus the baseline, while the \textcolor[HTML]{8dd3c7}{green lines} show the $\Delta\text{PSNR}$ of \textit{Ours} minus the baseline.
A larger $\Delta$ indicates greater performance improvement over the baseline. Thus, random cropping provides a slight improvement, while our method noticeably improves performance, particularly at higher exposure ratios.

\noindent \textbf{Realism validation of synthetic data.}
To further evaluate the realism of our synthetic noise, we train a denoising network using all available real dark frames. For a fair comparison, we generate synthetic dark frames with the same data volume and train another denoising network under identical settings. No performance degradation is observed, validating the realism of our synthetic data. Detailed experimental results are provided in the appendix.

\noindent \textbf{Photon noise parameter estimation with limited data.} 
As described in Sec.~\ref{sec:shot_noise_est}, the system gain $g$ can be estimated using varying amounts of data. In our method, we assume only a single noisy image per ISO. In this ablation, we assess the effect of data availability by comparing denoising performance on the SID and ELD test sets when $g$ is estimated using either one noisy image or 16 clean–noisy pairs. The experiment result (see appendix) shows that using a noisy image leads to a slight performance drop under some settings compared to using multiple clean–noisy pairs. 
For our main results (Table~\ref{tab:compare_sid} and Table~\ref{tab:compare_lrid}), we use the parameters estimated from a single noisy image per ISO. 
While in the above ablation studies, we use the parameters estimated from 16 clean-noisy pairs (fairness is ensured among all ablation studies).

\section{Limitations and Conclusion}
We addressed the challenge of obtaining large paired clean–noisy datasets for training low-light image denoisers by proposing a practical noise synthesis method that requires only one noisy image and one dark frame per ISO setting. We introduce a Fourier-domain spectral sampling strategy to model the complex signal-independent noise. This enables the generation of diverse and physically accurate noise realizations that preserve both the spectral characteristics and the marginal statistics of real sensor noise, without relying on simplified parametric models or large-scale data acquisition. The method is data-efficient, computationally lightweight, and can be applied to different sensors without extensive calibration.

Despite these advances, some limitations remain. Some components of signal-independent noise are influenced by factors beyond ISO, such as sensor temperature, exposure duration, and readout timing. For instance, dark current increases with temperature, and black level error may drift slightly as the sensor warms up. In our current formulation, we assume stable imaging conditions and model noise variations primarily with respect to ISO, without explicitly accounting for temperature or exposure time. Extending the method to incorporate these factors is an interesting direction for future work.

\bibliographystyle{plainnat}
\bibliography{main}

@String(CVPR= {IEEE Conf. Comput. Vis. Pattern Recog.})

@String(CVPR  = {CVPR})

@article{konnik2014high,
  title={High-level numerical simulations of noise in CCD and CMOS photosensors: review and tutorial},
  author={Konnik, Mikhail and Welsh, James},
  journal={arXiv preprint arXiv:1412.4031},
  year={2014}
}

@article{gow2007comprehensive,
  title={A comprehensive tool for modeling CMOS image-sensor-noise performance},
  author={Gow, Ryan D and Renshaw, David and Findlater, Keith and Grant, Lindsay and McLeod, Stuart J and Hart, John and Nicol, Robert L},
  journal={IEEE Transactions on Electron Devices},
  volume={54},
  number={6},
  pages={1321--1329},
  year={2007},
  publisher={IEEE}
}

@inproceedings{costantini2004virtual,
  title={Virtual sensor design},
  author={Costantini, Roberto and Susstrunk, Sabine},
  booktitle={Sensors and Camera Systems for Scientific, Industrial, and Digital Photography Applications V},
  volume={5301},
  pages={408--419},
  year={2004},
  organization={SPIE}
}

@article{wei2021physics,
  title={Physics-based noise modeling for extreme low-light photography},
  author={Wei, Kaixuan and Fu, Ying and Zheng, Yinqiang and Yang, Jiaolong},
  journal={IEEE Transactions on Pattern Analysis and Machine Intelligence},
  volume={44},
  number={11},
  pages={8520--8537},
  year={2021},
  publisher={IEEE}
}

@article{feng2023learnability,
  title={Learnability enhancement for low-light raw image denoising: A data perspective},
  author={Feng, Hansen and Wang, Lizhi and Wang, Yuzhi and Fan, Haoqiang and Huang, Hua},
  journal={IEEE Transactions on Pattern Analysis and Machine Intelligence},
  volume={46},
  number={1},
  pages={370--387},
  year={2023},
  publisher={IEEE}
}

@inproceedings{zhang2021rethinking,
  title={Rethinking noise synthesis and modeling in raw denoising},
  author={Zhang, Yi and Qin, Hongwei and Wang, Xiaogang and Li, Hongsheng},
  booktitle={Proceedings of the IEEE/CVF International Conference on Computer Vision},
  pages={4593--4601},
  year={2021}
}

@inproceedings{monakhova2022dancing,
  title={Dancing under the stars: video denoising in starlight},
  author={Monakhova, Kristina and Richter, Stephan R and Waller, Laura and Koltun, Vladlen},
  booktitle={Proceedings of the IEEE/CVF Conference on Computer Vision and Pattern Recognition},
  pages={16241--16251},
  year={2022}
}

@article{lu2025dark,
  title={Dark Noise Diffusion: Noise Synthesis for Low-Light Image Denoising},
  author={Lu, Liying and Achddou, Raphael and Susstrunk, Sabine},
  journal={IEEE Transactions on Pattern Analysis and Machine Intelligence},
  year={2025},
  publisher={IEEE}
}

@article{foi2008practical,
	title        = {Practical Poissonian-Gaussian noise modeling and fitting for single-image raw-data},
	author       = {Foi, Alessandro and Trimeche, Mejdi and Katkovnik, Vladimir and Egiazarian, Karen},
	year         = 2008,
	journal      = {IEEE Transactions on Image Processing},
	publisher    = {IEEE},
	volume       = 17,
	number       = 10,
	pages        = {1737--1754}
}

@article{foi2009clipped,
	title        = {Clipped noisy images: Heteroskedastic modeling and practical denoising},
	author       = {Foi, Alessandro},
	year         = 2009,
	journal      = {Signal Processing},
	publisher    = {Elsevier},
	volume       = 89,
	number       = 12,
	pages        = {2609--2629}
}

@inproceedings{brooks2019unprocessing,
	title        = {Unprocessing images for learned raw denoising},
	author       = {Brooks, Tim and Mildenhall, Ben and Xue, Tianfan and Chen, Jiawen and Sharlet, Dillon and Barron, Jonathan T},
	year         = 2019,
	booktitle    = {Proceedings of the IEEE/CVF Conference on Computer Vision and Pattern Recognition},
	pages        = {11036--11045}
}

@article{bahler2022pogain,
	title        = {Pogain: Poisson-gaussian image noise modeling from paired samples},
	author       = {B{\"a}hler, Nicolas and El Helou, Majed and Objois, {\'E}tienne and Okumu{\c{s}}, Kaan and S{\"u}sstrunk, Sabine},
	year         = 2022,
	journal      = {IEEE Signal Processing Letters},
	publisher    = {IEEE},
	volume       = 29,
	pages        = {2602--2606}
}

@article{joiner1971some,
  title={Some properties of the range in samples from Tukey's symmetric lambda distributions},
  author={Joiner, Brian L and Rosenblatt, Joan R},
  journal={Journal of the American Statistical Association},
  volume={66},
  number={334},
  pages={394--399},
  year={1971},
  publisher={Taylor \& Francis}
}

@inproceedings{zhang2023towards,
	title        = {Towards General Low-Light Raw Noise Synthesis and Modeling},
	author       = {Zhang, Feng and Xu, Bin and Li, Zhiqiang and Liu, Xinran and Lu, Qingbo and Gao, Changxin and Sang, Nong},
	year         = 2023,
	booktitle    = {Proceedings of the IEEE/CVF International Conference on Computer Vision},
	pages        = {10820--10830}
}

@inproceedings{abdelhamed2019noise,
	title        = {Noise flow: Noise modeling with conditional normalizing flows},
	author       = {Abdelhamed, Abdelrahman and Brubaker, Marcus A and Brown, Michael S},
	year         = 2019,
	booktitle    = {Proceedings of the IEEE/CVF International Conference on Computer Vision},
	pages        = {3165--3173}
}

@article{zhang2017improved,
	title        = {Improved denoising via Poisson mixture modeling of image sensor noise},
	author       = {Zhang, Jiachao and Hirakawa, Keigo},
	year         = 2017,
	journal      = {IEEE Transactions on Image Processing},
	publisher    = {IEEE},
	volume       = 26,
	number       = 4,
	pages        = {1565--1578}
}

@inproceedings{wach2004noise,
	title        = {Noise modeling for design and simulation of color imaging systems},
	author       = {Wach, Hans B and Dowski, Edward R},
	year         = 2004,
	booktitle    = {Color and Imaging Conference},
	volume       = 12,
	pages        = {211--216},
	organization = {Society of Imaging Science and Technology}
}

@book{nakamura2017image,
	title        = {Image sensors and signal processing for digital still cameras},
	author       = {Nakamura, Junichi},
	year         = 2017,
	publisher    = {CRC press}
}

@article{cain2001projection,
	title        = {Projection-based image registration in the presence of fixed-pattern noise},
	author       = {Cain, Stephen C and Hayat, Majeed M and Armstrong, Ernest E},
	year         = 2001,
	journal      = {IEEE Transactions on Image Processing},
	publisher    = {IEEE},
	volume       = 10,
	number       = 12,
	pages        = {1860--1872}
}

@book{holst2007cmos,
  title={CMOS/CCD Sensors},
  author={Holst, Gerald C and GC, Lomheim},
  year={2007},
  publisher={JCD publishing}
}

@article{dinh2016density,
	title        = {Density estimation using real nvp},
	author       = {Dinh, Laurent and Sohl-Dickstein, Jascha and Bengio, Samy},
	year         = 2016,
	journal      = {arXiv preprint arXiv:1605.08803}
}

@article{kingma2018glow,
	title        = {Glow: Generative flow with invertible 1x1 convolutions},
	author       = {Kingma, Durk P and Dhariwal, Prafulla},
	year         = 2018,
	journal      = {Advances in Neural Information Processing Systems},
	volume       = 31
}

@article{kobyzev2020normalizing,
	title        = {Normalizing flows: An introduction and review of current methods},
	author       = {Kobyzev, Ivan and Prince, Simon JD and Brubaker, Marcus A},
	year         = 2020,
	journal      = {IEEE Transactions on Pattern Analysis and Machine Intelligence},
	publisher    = {IEEE},
	volume       = 43,
	number       = 11,
	pages        = {3964--3979}
}

@inproceedings{kousha2022modeling,
	title        = {Modeling srgb camera noise with normalizing flows},
	author       = {Kousha, Shayan and Maleky, Ali and Brown, Michael S and Brubaker, Marcus A},
	year         = 2022,
	booktitle    = {Proceedings of the IEEE/CVF Conference on Computer Vision and Pattern Recognition},
	pages        = {17463--17471}
}

@article{gulrajani2017improved,
	title        = {Improved training of wasserstein gans},
	author       = {Gulrajani, Ishaan and Ahmed, Faruk and Arjovsky, Martin and Dumoulin, Vincent and Courville, Aaron C},
	year         = 2017,
	journal      = {Advances in Neural Information Processing Systems},
	volume       = 30
}

@inproceedings{mescheder2018training,
	title        = {Which training methods for GANs do actually converge?},
	author       = {Mescheder, Lars and Geiger, Andreas and Nowozin, Sebastian},
	year         = 2018,
	booktitle    = {International Conference on Machine Learning},
	pages        = {3481--3490},
	organization = {PMLR}
}

@article{ramesh2022hierarchical,
	title        = {Hierarchical text-conditional image generation with clip latents},
	author       = {Ramesh, Aditya and Dhariwal, Prafulla and Nichol, Alex and Chu, Casey and Chen, Mark},
	year         = 2022,
	journal      = {arXiv preprint arXiv:2204.06125},
	volume       = 1,
	number       = 2,
	pages        = 3
}

@inproceedings{rombach2022high,
	title        = {High-resolution image synthesis with latent diffusion models},
	author       = {Rombach, Robin and Blattmann, Andreas and Lorenz, Dominik and Esser, Patrick and Ommer, Bj{\"o}rn},
	year         = 2022,
	booktitle    = {Proceedings of the IEEE/CVF Conference on Computer Vision and Pattern Recognition},
	pages        = {10684--10695}
}

@article{galerne2010random,
  title={Random phase textures: Theory and synthesis},
  author={Galerne, Bruno and Gousseau, Yann and Morel, Jean-Michel},
  journal={IEEE Transactions on image processing},
  volume={20},
  number={1},
  pages={257--267},
  year={2010},
  publisher={IEEE}
}

@article{healey2002radiometric,
  title={Radiometric CCD camera calibration and noise estimation},
  author={Healey, Glenn E and Kondepudy, Raghava},
  journal={IEEE Transactions on Pattern Analysis and Machine Intelligence},
  volume={16},
  number={3},
  pages={267--276},
  year={2002},
  publisher={IEEE}
}

@inproceedings{cao2023physics,
  title={Physics-guided iso-dependent sensor noise modeling for extreme low-light photography},
  author={Cao, Yue and Liu, Ming and Liu, Shuai and Wang, Xiaotao and Lei, Lei and Zuo, Wangmeng},
  booktitle={Proceedings of the IEEE/CVF Conference on Computer Vision and Pattern Recognition},
  pages={5744--5753},
  year={2023}
}

@inproceedings{chen2018learning,
  title={Learning to see in the dark},
  author={Chen, Chen and Chen, Qifeng and Xu, Jia and Koltun, Vladlen},
  booktitle={Proceedings of the IEEE conference on computer vision and pattern recognition},
  pages={3291--3300},
  year={2018}
}

@inproceedings{lewis1984texture,
  title={Texture synthesis for digital painting},
  author={Lewis, John-Peter},
  booktitle={Proceedings of the 11th annual conference on Computer graphics and interactive techniques},
  pages={245--252},
  year={1984}
}

@inproceedings{van1991spot,
  title={Spot noise texture synthesis for data visualization},
  author={Van Wijk, Jarke J},
  booktitle={Proceedings of the 18th annual conference on Computer graphics and interactive techniques},
  pages={309--318},
  year={1991}
}

@inproceedings{abdelhamed2018high,
  title={A high-quality denoising dataset for smartphone cameras},
  author={Abdelhamed, Abdelrahman and Lin, Stephen and Brown, Michael S},
  booktitle={Proceedings of the IEEE conference on computer vision and pattern recognition},
  pages={1692--1700},
  year={2018}
}

@article{anaya2018renoir,
  title={Renoir--a dataset for real low-light image noise reduction},
  author={Anaya, Josue and Barbu, Adrian},
  journal={Journal of Visual Communication and Image Representation},
  volume={51},
  pages={144--154},
  year={2018},
  publisher={Elsevier}
}

@inproceedings{flepp2024real,
  title={Real-world mobile image denoising dataset with efficient baselines},
  author={Flepp, Roman and Ignatov, Andrey and Timofte, Radu and Van Gool, Luc},
  booktitle={Proceedings of the IEEE/CVF Conference on Computer Vision and Pattern Recognition},
  pages={22368--22377},
  year={2024}
}

@article{zhang2017beyond,
  title={Beyond a gaussian denoiser: Residual learning of deep cnn for image denoising},
  author={Zhang, Kai and Zuo, Wangmeng and Chen, Yunjin and Meng, Deyu and Zhang, Lei},
  journal={IEEE transactions on image processing},
  volume={26},
  number={7},
  pages={3142--3155},
  year={2017},
  publisher={IEEE}
}

@inproceedings{zamir2022restormer,
  title={Restormer: Efficient transformer for high-resolution image restoration},
  author={Zamir, Syed Waqas and Arora, Aditya and Khan, Salman and Hayat, Munawar and Khan, Fahad Shahbaz and Yang, Ming-Hsuan},
  booktitle={Proceedings of the IEEE/CVF conference on computer vision and pattern recognition},
  pages={5728--5739},
  year={2022}
}

@inproceedings{liang2021swinir,
  title={Swinir: Image restoration using swin transformer},
  author={Liang, Jingyun and Cao, Jiezhang and Sun, Guolei and Zhang, Kai and Van Gool, Luc and Timofte, Radu},
  booktitle={Proceedings of the IEEE/CVF international conference on computer vision},
  pages={1833--1844},
  year={2021}
}

@article{mao2016image,
  title={Image restoration using very deep convolutional encoder-decoder networks with symmetric skip connections},
  author={Mao, Xiaojiao and Shen, Chunhua and Yang, Yu-Bin},
  journal={Advances in neural information processing systems},
  volume={29},
  year={2016}
}

@inproceedings{buades2005non,
  title={A non-local algorithm for image denoising},
  author={Buades, Antoni and Coll, Bartomeu and Morel, J-M},
  booktitle={2005 IEEE computer society conference on computer vision and pattern recognition (CVPR'05)},
  volume={2},
  pages={60--65},
  year={2005},
  organization={Ieee}
}

@inproceedings{gu2014weighted,
  title={Weighted nuclear norm minimization with application to image denoising},
  author={Gu, Shuhang and Zhang, Lei and Zuo, Wangmeng and Feng, Xiangchu},
  booktitle={Proceedings of the IEEE conference on computer vision and pattern recognition},
  pages={2862--2869},
  year={2014}
}

@article{zhang2018ffdnet,
  title={FFDNet: Toward a fast and flexible solution for CNN-based image denoising},
  author={Zhang, Kai and Zuo, Wangmeng and Zhang, Lei},
  journal={IEEE Transactions on Image Processing},
  volume={27},
  number={9},
  pages={4608--4622},
  year={2018},
  publisher={IEEE}
}

@article{zhang2021plug,
  title={Plug-and-play image restoration with deep denoiser prior},
  author={Zhang, Kai and Li, Yawei and Zuo, Wangmeng and Zhang, Lei and Van Gool, Luc and Timofte, Radu},
  journal={IEEE Transactions on Pattern Analysis and Machine Intelligence},
  volume={44},
  number={10},
  pages={6360--6376},
  year={2021},
  publisher={IEEE}
}

\appendix
\clearpage
\appendix
\section*{Appendix}

This appendix is organized as follows:

\begin{itemize}
    \item Method Details $\rightarrow$ Sec.~\ref{supp:method_details}
    \begin{itemize}
        \item Spectral Sampling Algorithm $\rightarrow$ Sec.~\ref{supp:spectral_sampling}
        \item Photon Noise Parameter Estimation $\rightarrow$ Sec.~\ref{supp:photon_noise_est}
    \end{itemize}
    
    \item More Implementation Details $\rightarrow$ Sec.~\ref{supp:implementation}
    \begin{itemize}
        \item Data Synthesis Details $\rightarrow$ Sec.~\ref{supp:data_details}
        \item Training Details $\rightarrow$ Sec.~\ref{supp:traininig_details}
        \item Evaluation Details $\rightarrow$ Sec.~\ref{supp:evaluation_details}
    \end{itemize}
     \item More Discussions $\rightarrow$ Sec.~\ref{supp:more_dis}
    \begin{itemize}
        \item Effect of the Fixed-Pattern Removal Before Sampling $\rightarrow$ Sec.~\ref{supp:fix_removal}
        \item Importance of Inter-Channel Correlation $\rightarrow$ Sec.~\ref{supp:icc}
        \item Effect of Iterative Histogram Matching $\rightarrow$ Sec.~\ref{supp:ihm}
        \item Different Ways to Leverage a Single Dark Frame $\rightarrow$ Sec.~\ref{supp:leverage_single}
        \item Realism Validation of Synthetic Data $\rightarrow$ Sec.~\ref{supp:realism}
        \item  Photon Noise Parameter Estimation with Limited Data $\rightarrow$ Sec.~\ref{supp:photon_limited}
    \end{itemize}
    \item  Synthetic Noise Visualization Across Different Sensors  $\rightarrow$ Sec.~\ref{supp:noise_vis}
    \item More Visual Comparisons $\rightarrow$ Sec.~\ref{supp:more_visual}
\end{itemize}

\vspace{20pt}

For optimal visual comparisons, we recommend viewing the results on a screen. 

\vspace{20pt}

\section{Method Details}
\label{supp:method_details}

\subsection{Spectral Sampling Algorithm}
\label{supp:spectral_sampling}

The full procedure of our spectral sampling algorithm for dark-frame synthesis is summarized in Algorithm~\ref{alg:spectral_sampling}.

\begin{algorithm}[t]
\caption{Spectral Sampling with Iterative Refinement for Dark Frame Synthesis}
\label{alg:spectral_sampling}
\SetAlgoLined
\KwIn{Reference dark frame $I_{\text{dark}} \in \mathbb{R}^{C \times H \times W}$, Gaussian blur parameter $\sigma$, number of iterations $K$}
\KwOut{Synthesized dark frame $\tilde{I}_{\text{dark}}$}

\textbf{Step 1: Fixed-pattern removal}\\
\Indp 
$S \leftarrow \mathcal{G}_{\sigma}(I_{\text{dark}})$ \tcp*{Extract low-frequency pattern}
$R' \leftarrow I_{\text{dark}} - S$ \tcp*{Isolate stochastic residual}
$\mu^{R} \leftarrow \mathbb{E}[R']$ \tcp*{Compute channel-wise mean}
$R \leftarrow R' - \mu^{R}$ \tcp*{Center to zero mean}
\Indm

\textbf{Step 2: Spectral prior estimation}\\
\Indp 
$\widehat{R} \leftarrow \mathcal{F}\{R\}$ \tcp*{Compute reference spectrum}
\Indm

\textbf{Step 3: Phase randomization}\\
\Indp 
$\xi^0 \sim \mathcal{U} \big( [-\pi, \pi]^{(H,W)} \big)$ \tcp*{Uniform random phase}
$\xi \leftarrow \text{replicate}(\xi^0, C)$ \tcp*{Copy across channels}
$\widehat{N} \leftarrow |\widehat{R}| \odot \exp\Big(i \big(\theta_{\widehat{R}} + \xi\big)\Big)$ \tcp*{Phase Randomization}
$N^{(0)} \leftarrow \frac{1}{\sqrt{HW}}\mathcal{F}^{-1}\{\widehat{N}\}$ \tcp*{Initial synthesis}
\Indm

\textbf{Step 4: Iterative histogram-spectral refinement}\\
\Indp 
\For{$k = 0$ \KwTo $K-1$}{
    $N'^{(k)}_{\text{hist}} \leftarrow \mathcal{H}(N^{(k)}, R)$ \tcp*{Match Hist.}

    $\mu^{(k)}_{hist} \leftarrow \mathbb{E}[N'^{(k)}_{\text{hist}}]$ \\
    $N^{(k)}_{\text{hist}} \leftarrow N'^{(k)}_{\text{hist}} - \mu^{(k)}_{hist}$ \\
    
    $\widehat{N}^{(k)}_{\text{hist}} \leftarrow \mathcal{F}\{N^{(k)}_{\text{hist}}\}$ \\
    $\widehat{N}^{(k)}_{\text{corrected}} \leftarrow |\widehat{R}| \odot \exp(i \theta_{\widehat{N}^{(k)}_{\text{hist}}})$ \tcp*{Restore spectrum}
    $N^{(k+1)} \leftarrow \frac{1}{\sqrt{HW}}\mathcal{F}^{-1}\{\widehat{N}^{(k)}_{\text{corrected}}\} + \mu^{(k)}_{hist}$ \\
}
\Indm

\textbf{Step 5: Final reconstruction}\\
\Indp 
$\tilde{I}_{\text{dark}} \leftarrow N^{(K)} + S + \mu^{R}$ \tcp*{Restore fixed pattern and mean}
\Indm

\Return $\tilde{I}_{\text{dark}}$

\end{algorithm}


\subsection{Photon Noise Parameter Estimation}
\label{supp:photon_noise_est}

In Sec.~\ref{sec:shot_noise_est} of the main paper, we have shown that the noisy observation $y$ can be modeled as: 
\begin{align}
    y = g\mathcal{P}(x) + n_{\text{other}}.
\end{align}

For a Poisson random variable, the variance equals its mean, i.e., $\text{Var}(\mathcal{P}(x)) = x$. Thus the variance of the noisy observation is:
\begin{align}
\text{Var}(y) &= g^2 \text{Var}(\mathcal{P}(x)) + \text{Var}(n_{\text{other}}) \nonumber \\
&= g^2 x + \text{Var}(n_{\text{other}}) \nonumber \\
&= g(gx) + \text{Var}(n_{\text{other}}).
\end{align}

Consequently, the system gain $g$ can be estimated by performing a linear fit between the variance of the noisy observation $y$ and the amplified clean signal $gx$, using a set of clean–noisy pairs. If only a single noisy image is available, a rough estimate of $g$ can still be obtained.
Specifically, we unfold the noisy image into overlapping $3\times3$ patches. For each patch, we apply a Gaussian blur to obtain a pseudo-clean value, and all pixels within the patch serve as noisy observations from which the noise variance is computed.
Patches sharing the same mean value are then grouped together to produce more robust variance estimates.

Of course, more sophisticated approaches can be used for pseudo-clean estimation when needed, and additional data can possibly further improve accuracy. For example, when flat-field frames are available (\eg, a uniform calibration card illuminated by a single light source), a more reliable estimation of $g$ can be obtained by following the procedures described in \cite{wei2021physics,feng2023learnability,healey2002radiometric}.


\section{More Implementation Details}
\label{supp:implementation}

\subsection{Data Synthesis Details}
\label{supp:data_details}

In the LRID dataset~\cite{feng2023learnability}, two types of dark frames of ISO=6400 are provided: one captured under normal sensor temperature and the other in the sensor’s hot mode. For noise synthesis, we use a single dark frame from each setting to generate the corresponding synthetic dark frames. The synthesis results are shown in Fig.~\ref{fig:full_darkframe_visualization_lrid}.

\subsection{Training Details}
\label{supp:traininig_details}

We use the same U\textendash Net architecture as \cite{chen2018learning} for all denoising experiments. 
The networks are trained for 500 epochs using the Adam optimizer. 
The initial learning rate is set to $2\times 10^{-4}$, reduced by half at epoch 250, and fixed to $1\times 10^{-5}$ after epoch 400. 
During training, we use a batch size of 4 and randomly crop $256\times256$ patches with random horizontal flips for data augmentation. 
We train the denoisers using the $L_1$ loss between the network output and the clean reference image.

\subsection{Evaluation Details}
\label{supp:evaluation_details}

Since Peak Signal-to-Noise Ratio (PSNR) and Structural SIMilarity (SSIM) are sensitive to global illumination, even small brightness shifts can cause large metric variations. 
 During denoising inference, we apply the illumination correction method from the ELD paper~\cite{wei2021physics} to the denoised images. We apply this illumination correction to all compared methods before evaluating PSNR and SSIM.  We compute the PSNR and SSIM between the denoised and reference images in the RAW domain.


\section{More Discussions}
\label{supp:more_dis}

\subsection{Effect of the Fixed-Pattern Removal Before Sampling}
\label{supp:fix_removal}

Dark frames contain both deterministic fixed patterns and stochastic noise components. 
Directly applying spectral analysis to raw dark frames would conflate these two components: the low-frequency fixed patterns would dominate the power spectrum, biasing the estimated noise statistics and leading to synthesized noise that incorrectly replicates deterministic sensor artifacts as if they were stochastic variations.

\begin{figure}[t]
\centering
\includegraphics[width=\linewidth]{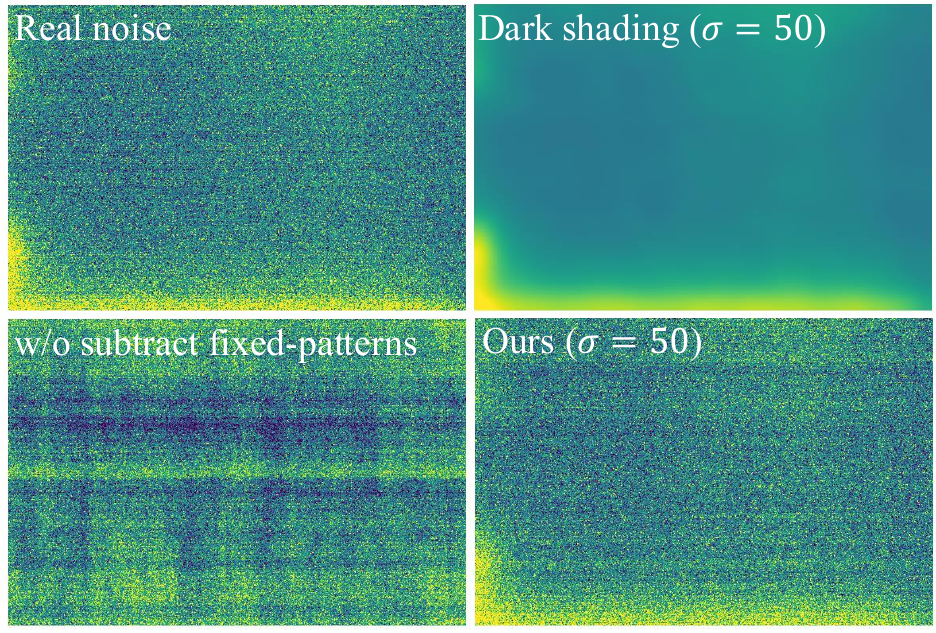}
\caption{Effect of fixed-pattern removal before noise synthesis. Left: Real dark frame at ISO 12800 (top) and synthesized noise without fixed-pattern removal (bottom), which incorrectly reproduces deterministic row/column structure.
Right: The estimated fixed-pattern component (top) and the synthesized noise after removing it prior to spectral sampling (bottom), resulting in a more realistic stochastic noise pattern.
}
\label{fig:fixed_pattern_removal}
\end{figure}

To isolate the stochastic component for spectral modeling, we apply Gaussian blur (Eq.~\ref{eq:gaussian_blur}) to extract the fixed pattern, then work with the residual. 
Figure~\ref{fig:fixed_pattern_removal} demonstrates the effect of fixed-pattern removal for noise synthesis. Given a real dark frame, without fixed-pattern removal, the synthesized noise (bottom-left) incorrectly produces random patterns of the deterministic fixed pattern, failing to produce faithful results. With appropriate blur kernel sizes, the method successfully separates the two components. We empirically find that choosing $\sigma$ in the range of $[40, 60]$ consistently yields good results. We set $\sigma=50$ in our experiments.


\subsection{Importance of Inter-Channel Correlation}
\label{supp:icc}

\begin{figure*}[t]
\centering
\includegraphics[width=\linewidth]{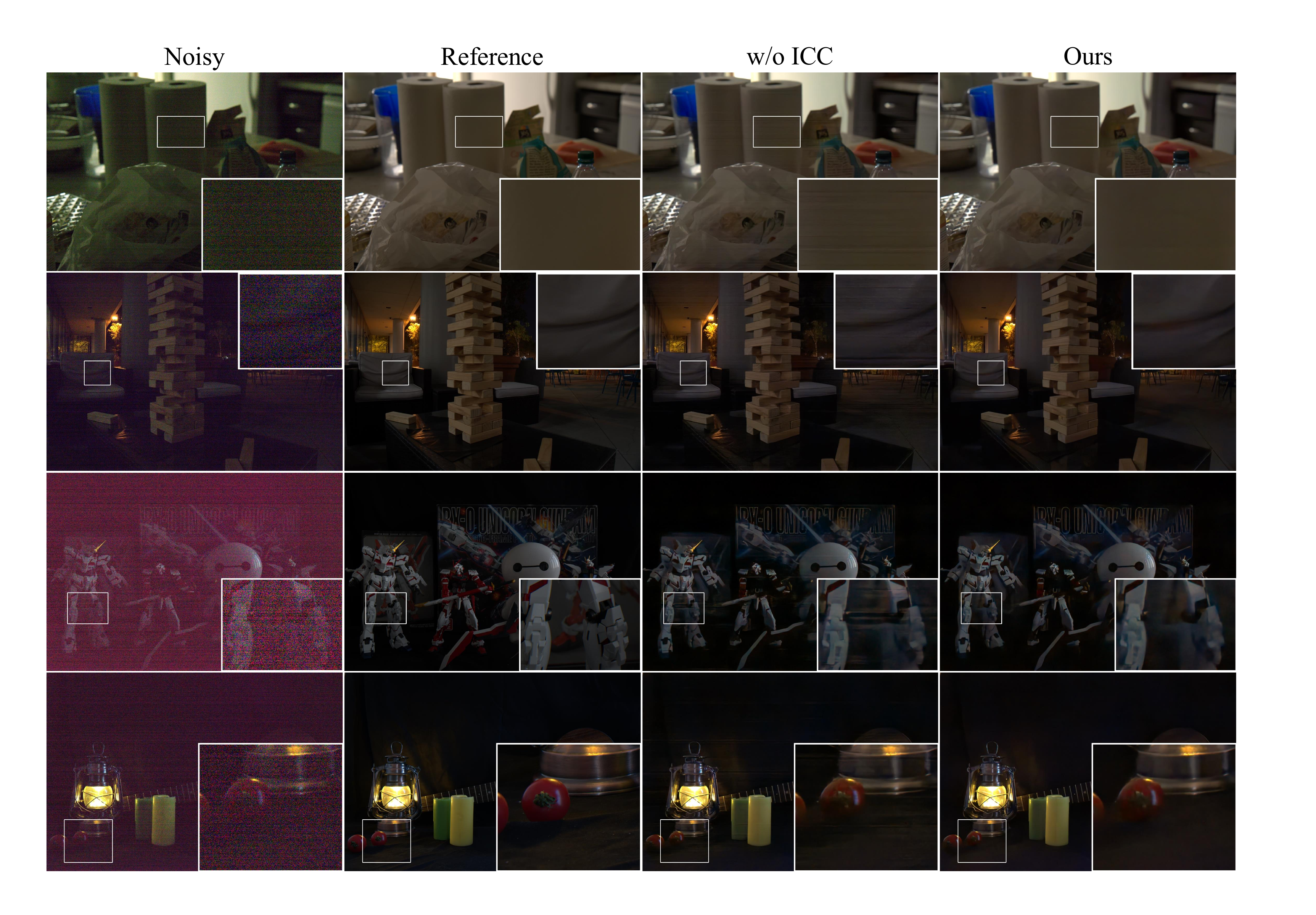}
\caption{Ablation study on inter-channel correlation (ICC). Without ICC, the denoised results show residual banding artifacts, while our full method produces clean outputs. The first two examples are from the SID test set, and the third and fourth are from the ELD test set. Best viewed when zoomed in.}
\label{fig:ablation_banding_appendix}
\end{figure*}

Fig.~\ref{fig:ablation_banding_appendix} presents additional ablation results for the inter-channel correlation~ICC. Without ICC, the denoised images exhibit noticeable residual banding artifacts, whereas our full method produces cleaner results.

\subsection{Effect of Iterative Histogram Matching}
\label{supp:ihm}

\begin{figure}[t]
\centering
\includegraphics[width=\linewidth]{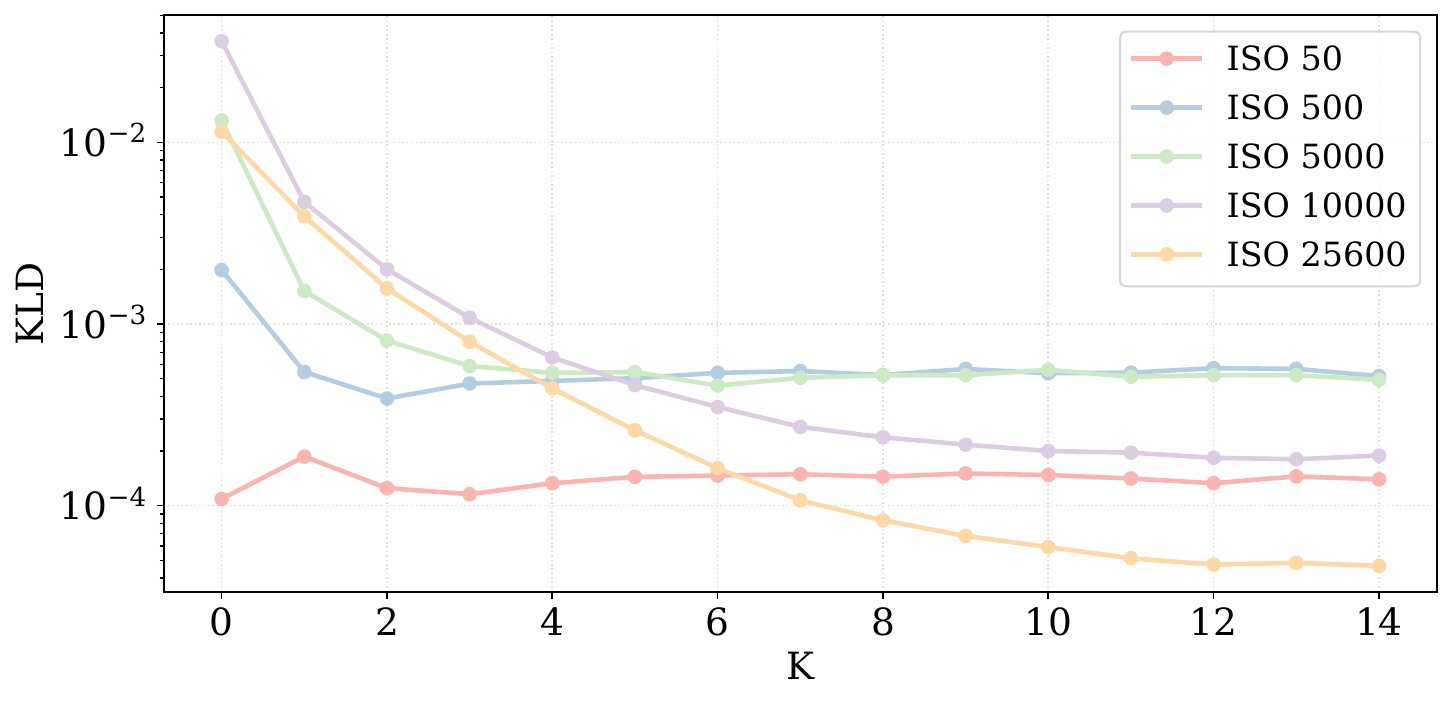}
\caption{Relationship between the Kullback–Leibler divergence (KLD) between the synthetic and real noise and the histogram matching iteration number $K$, across different ISO settings. Increasing $K$ generally reduces the KLD.
}
\label{fig:kld_plot}
\end{figure}

In the main paper, we mentioned that we perform $K$ iterations of histogram matching. Here, we study how the choice of $K$ influences the accuracy of the synthesized noise. In Fig.~\ref{fig:kld_plot}, we show the relationship between the Kullback–Leibler divergence (KLD) between the synthetic and real noise and the value of $K$. As $K$ increases, the synthetic noise progressively converges toward the real noise distribution, reflected by the decreasing KLD. We set $K = 10$ in our experiments.

\begin{figure*}[t]
\centering
\includegraphics[width=\linewidth]{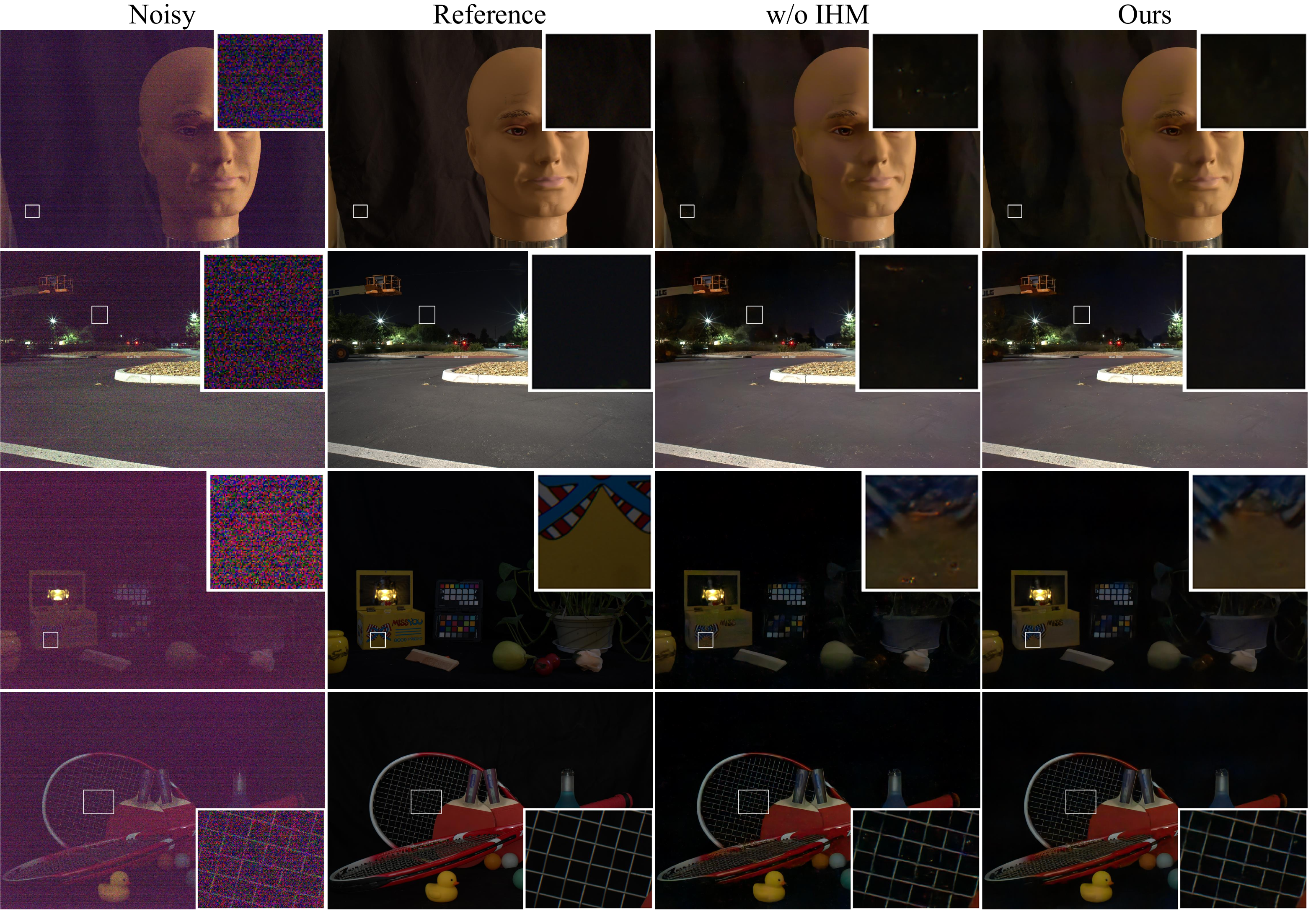}
\caption{Ablation study on the iterative histogram matching (IHM). Without IHM, noticeable white spots appear in the denoised outputs due to saturated pixels in the noisy inputs. With our full method, this issue is effectively mitigated. The first two examples are from the SID test set while the third and forth are from the ELD test set. Best viewed zoomed in.
}
\label{fig:saturated_pixels}
\end{figure*}

In the main paper, we showed that iterative histogram matching~(IHM) reduces color distortions in denoised results. Beyond this, we also find that IHM helps suppress saturated malfunctioning pixels. As shown in Fig.~\ref{fig:saturated_pixels}, without IHM, noticeable white spots appear in the denoised outputs due to saturated pixels in the noisy inputs. With our full method, this issue is effectively mitigated.

\begin{figure}[t]
\centering
\includegraphics[width=1.0\columnwidth]{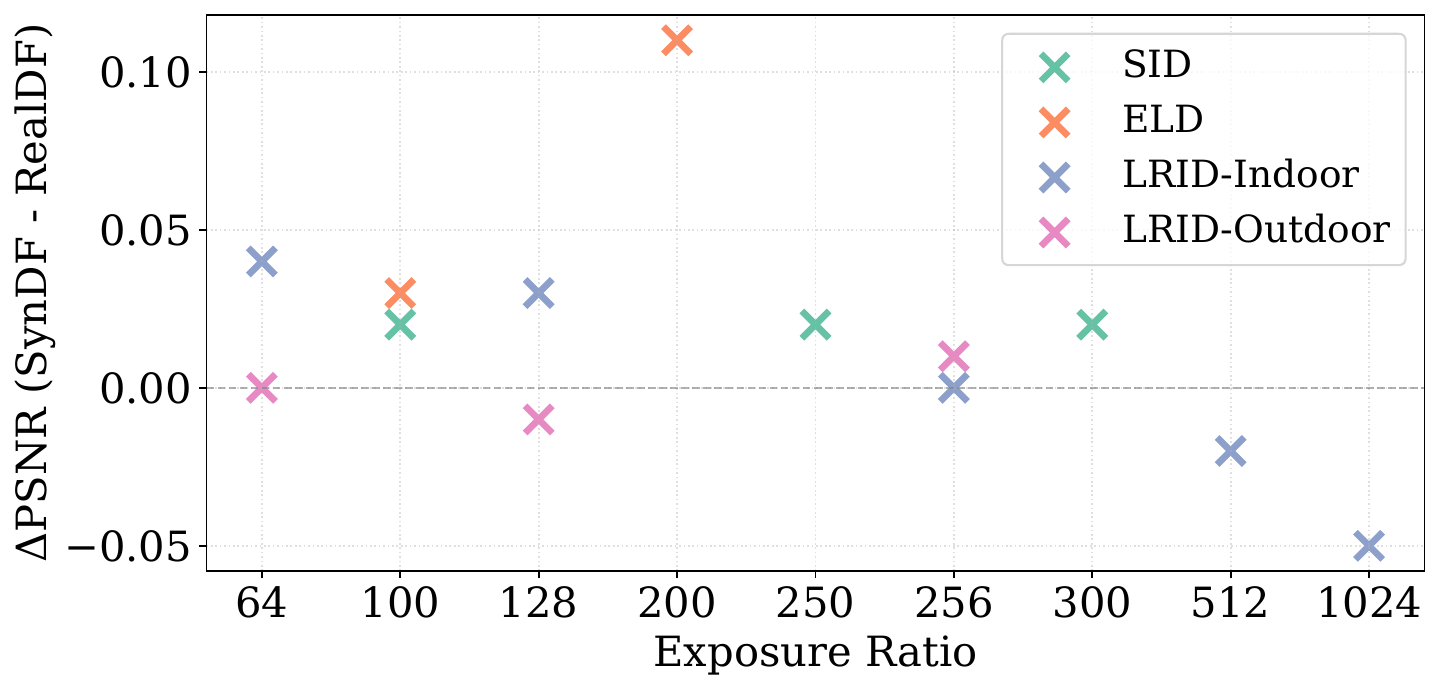}
\caption{PSNR differences between \textit{RealDF} and \textit{SynDF} across different exposure ratios for the SID, ELD, and LRID test sets. Each $\bm{\times}$ denotes the result of a subset of each test set, e.g., the left-most 
\textcolor[HTML]{DA8FC0}{$\bm{\times}$} denotes that the PSNR difference between \textit{RealDF} and \textit{SynDF} on the LRID-Indoor dataset with exposure Ratio = 64 is almost 0, indicating that our method can synthesize noise distributions nearly indistinguishable from real sensor noise.}
\label{fig:realism}
\end{figure}

\begin{figure*}[t]
\centering
\includegraphics[width=\linewidth]{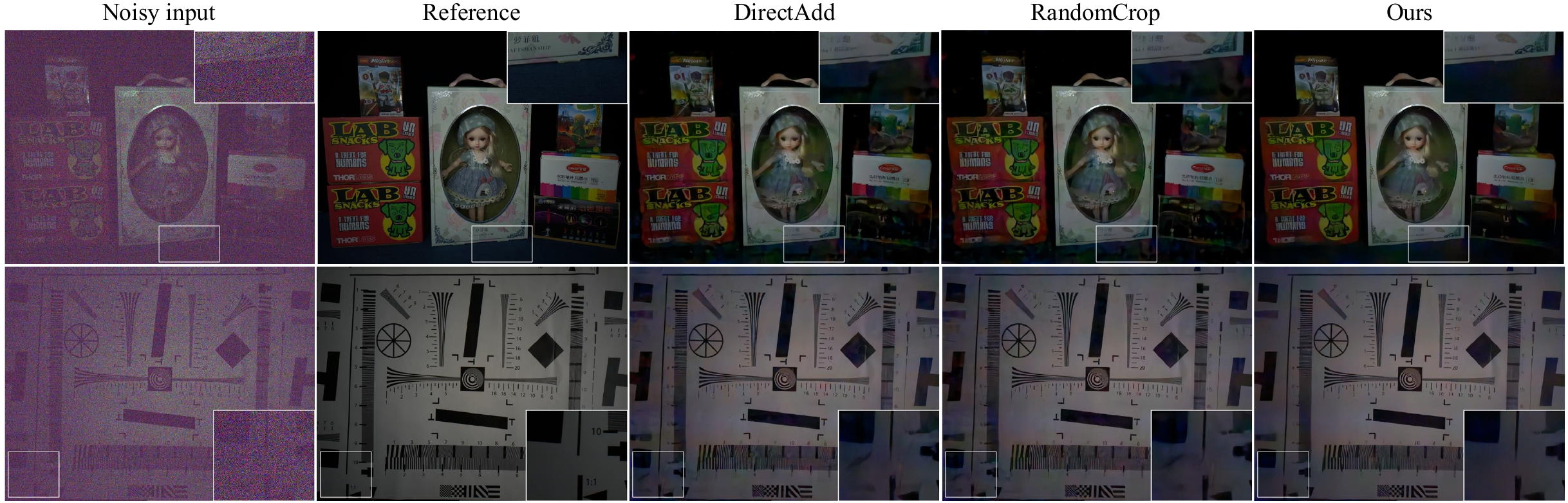}
\caption{Denoising results of the network trained with different data sources synthesized using only one real dark frame per ISO setting. Our method produces cleaner results with less artifacts.}
\label{fig:use_one_real}
\end{figure*}

\subsection{Different Ways to Leverage a Single Dark Frame}
\label{supp:leverage_single}

Given one single dark frame, we study the three different ways to leverage it for data synthesis: \textit{DirectAdd}, \textit{RandomCrop}, and \textit{Ours}. In Fig.~\ref{fig:use_one_real}, we show two denoising examples, our method produces cleaner results with fewer artifacts.

\subsection{Realism Validation of Synthetic Data}
\label{supp:realism}

To further evaluate the realism of our synthetic noise, we train a denoising network (\textit{RealDF}) using all available real dark frames from each ISO level in the SID dataset. For fair comparison, we generate one synthetic dark frame per real dark frame, ensuring identical data volume and ISO distribution, and train another denoising network (\textit{SynDF}). The same protocol is applied to the LRID dataset. 
Fig.~\ref{fig:realism} shows the PSNR difference between \textit{RealDF} and \textit{SynDF} on the SID, ELD and LRID test set across different exposure rations. 
The differences are consistently close to zero across all datasets and ratios, indicating that our method can synthesize noise distributions nearly indistinguishable from real sensor noise.

\subsection{Photon Noise Parameter Estimation with Limited Data} 
\label{supp:photon_limited}

In our experiments, we assume only a single noisy image per ISO is available for system gain estimation. To assess the effect of data availability, we compare denoising performance on the SID and ELD test sets when $g$ is estimated using either one noisy image or 16 clean–noisy pairs.

As shown in Table~\ref{tab:shot_param_est}, using a single noisy image leads to a slight drop in performance under some settings compared to using multiple clean–noisy pairs. While our study focuses on the single noisy image scenario, access to additional data, such as multiple clean–noisy pairs or flat-field frames, could potentially improve the accuracy of gain estimation and denoising results.

\begin{table}[t]
    \caption{Denoising performance when the system gain $g$ is estimated from either a single noisy image or 16 clean–noisy pairs. Results are reported in PSNR / SSIM.}
    \centering
    \resizebox{0.8\columnwidth}{!}{
    \begin{tabular}{ccccc}
    \toprule
    \textbf{Dataset}  & \textbf{Ratio} & 16 pairs & 1 noisy \\ 
    \midrule
    \multirow{3}{*}{\textbf{SID}} 
      & $\times$100 & 43.72 / 0.961 & 43.57 / 0.961  \\ 
      & $\times$250 & 41.30 / 0.944 & 41.24 / 0.945  \\ 
      & $\times$300 & 37.86 / 0.929 & 37.77 / 0.929  \\ 
      
    \cdashline{1-4}
    \multirow{2}{*}{\textbf{ELD}} 
      & $\times$100 & 47.14 / 0.986 & 47.13 / 0.986  \\ 
      & $\times$200 & 44.78 / 0.966 & 44.89 / 0.969  \\ 

    \bottomrule 
    \end{tabular}}
    \label{tab:shot_param_est}
\end{table}

\section{Synthetic Noise Visualization Across Different Sensors}
\label{supp:noise_vis}

\begin{figure*}[t]
\centering
\includegraphics[width=\linewidth]{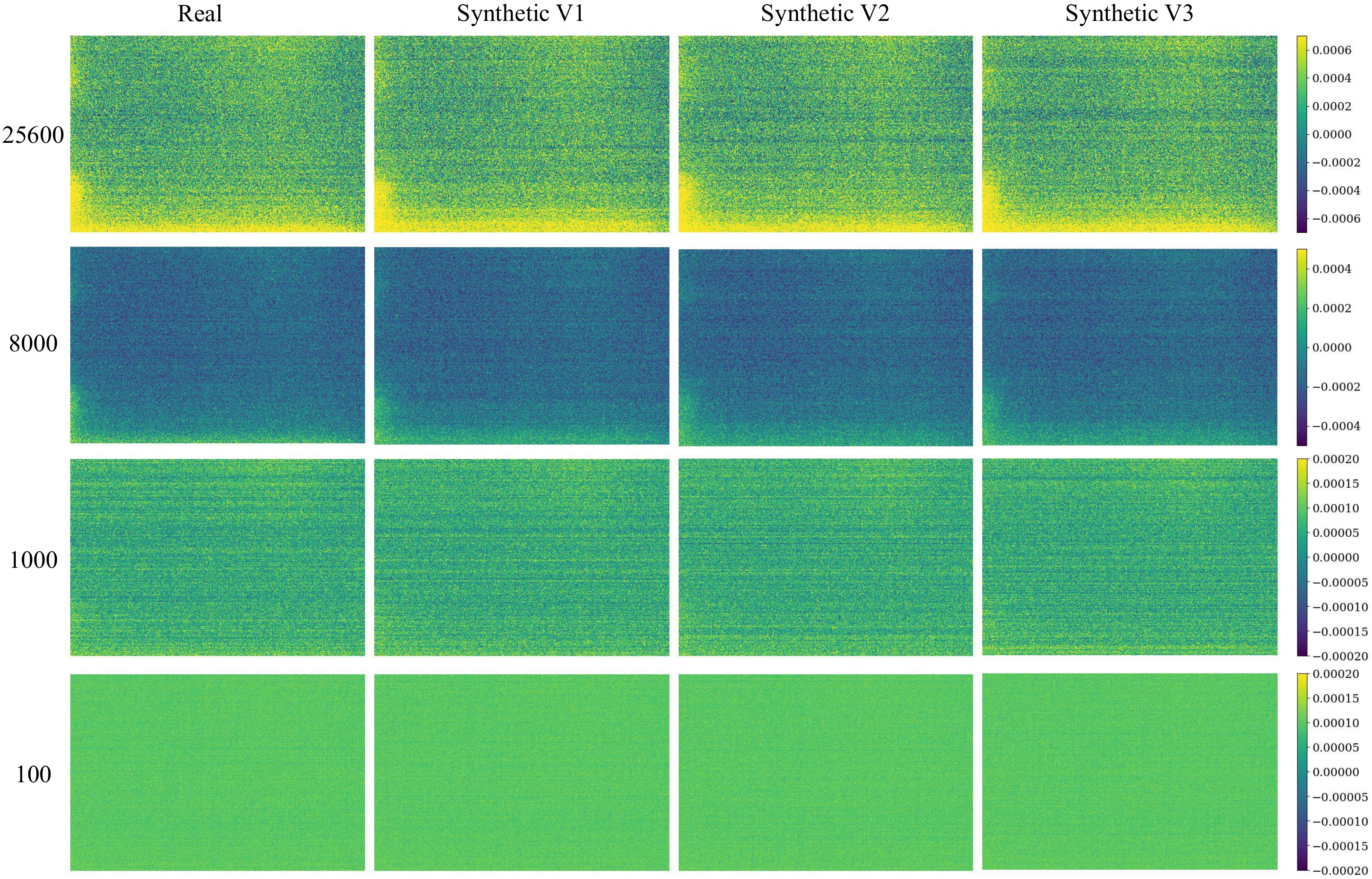}
\caption{Real dark frames and three different synthetic dark frame realizations (generated with different random seeds) across various ISO settings (ISO=25600, 8000, 1000, 100) of the Sony A7S2 sensor.}
\label{fig:full_darkframe_visualization_sid}
\end{figure*}

\begin{figure*}[t]
\centering
\includegraphics[width=\linewidth]{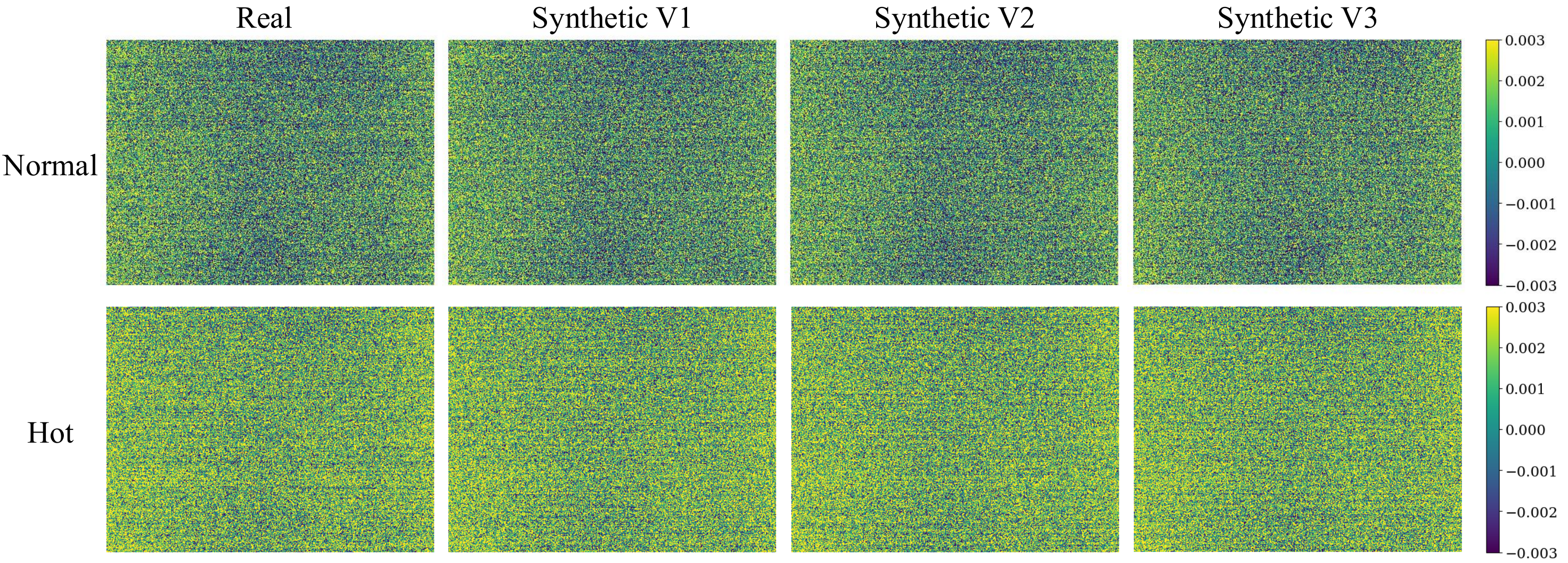}
\caption{Real dark frames and three different synthetic dark frame realizations (generated with different random seeds) with ISO=6400 of the IMX686 sensor. The first example is captured under normal sensor temperature and the second in the sensor’s hot mode.}
\label{fig:full_darkframe_visualization_lrid}
\end{figure*}

\begin{figure*}[t]
\centering
\includegraphics[width=\linewidth]{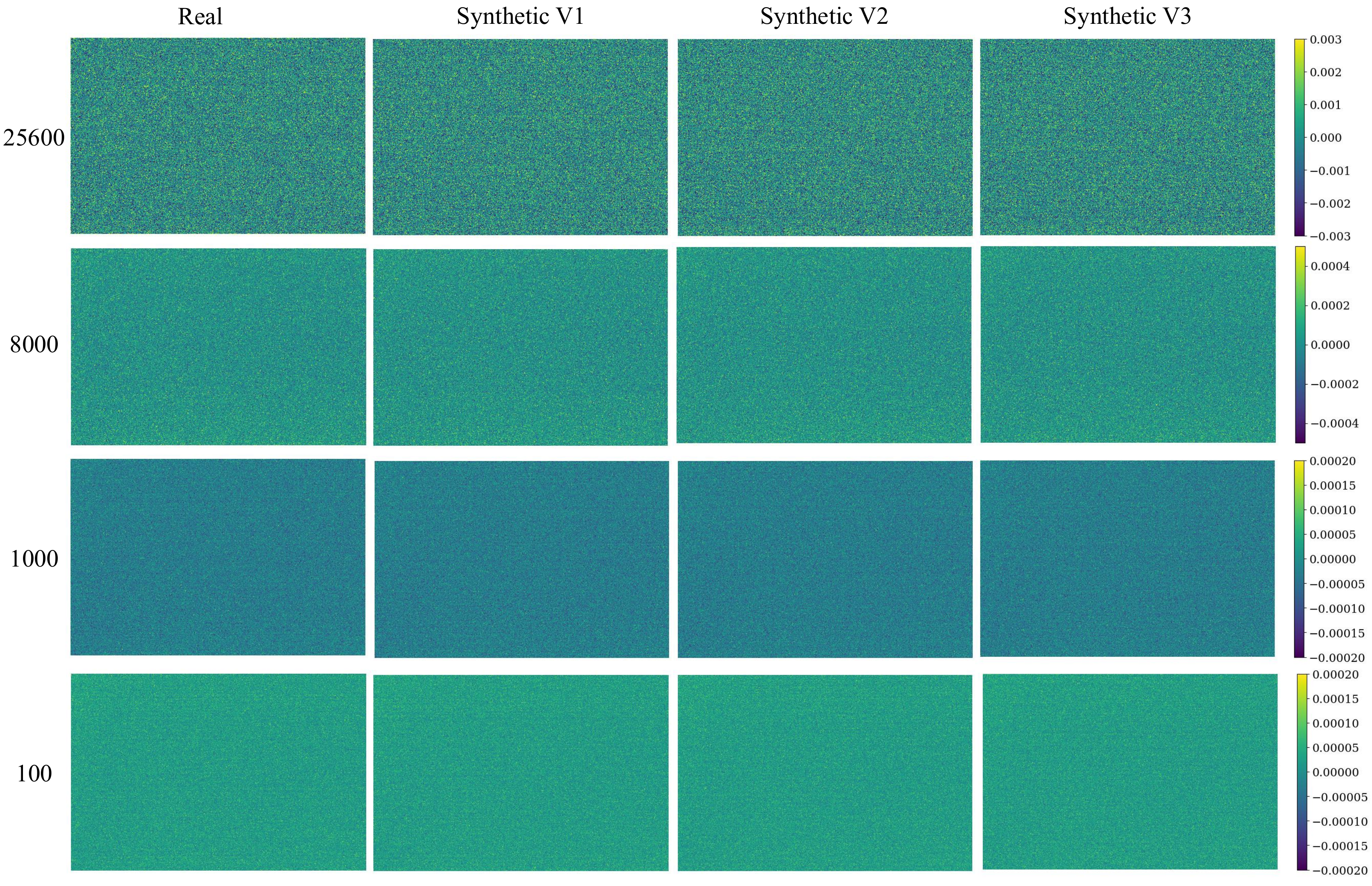}
\caption{Real dark frames and three different synthetic dark frame realizations (generated with different random seeds) across various ISO settings (ISO=25600, 8000, 1000, 100) of the Fujifilm X-M5 sensor.}
\label{fig:full_darkframe_visualization_fuji}
\end{figure*}

In this section, we visualize examples of real dark frames from various sensors alongside our synthesized dark frames. Figures~\ref{fig:full_darkframe_visualization_sid} and \ref{fig:full_darkframe_visualization_lrid} show examples from the Sony A7S2 sensor and the IMX686 sensor in the Redmi K30 smartphone, respectively. 
We also captured dark frames using a Fujifilm X-M5 camera, whose RAW images follow the X-Trans pattern instead of the Bayer pattern. Fig.~\ref{fig:full_darkframe_visualization_fuji} presents the real dark frames and our synthetic results. It can be observed that our method successfully generates realistic dark frames across different sensors, demonstrating its generalizability.

\section{More Visual Comparisons}
\label{supp:more_visual}

\begin{figure*}[t]
\centering
\includegraphics[width=\linewidth]{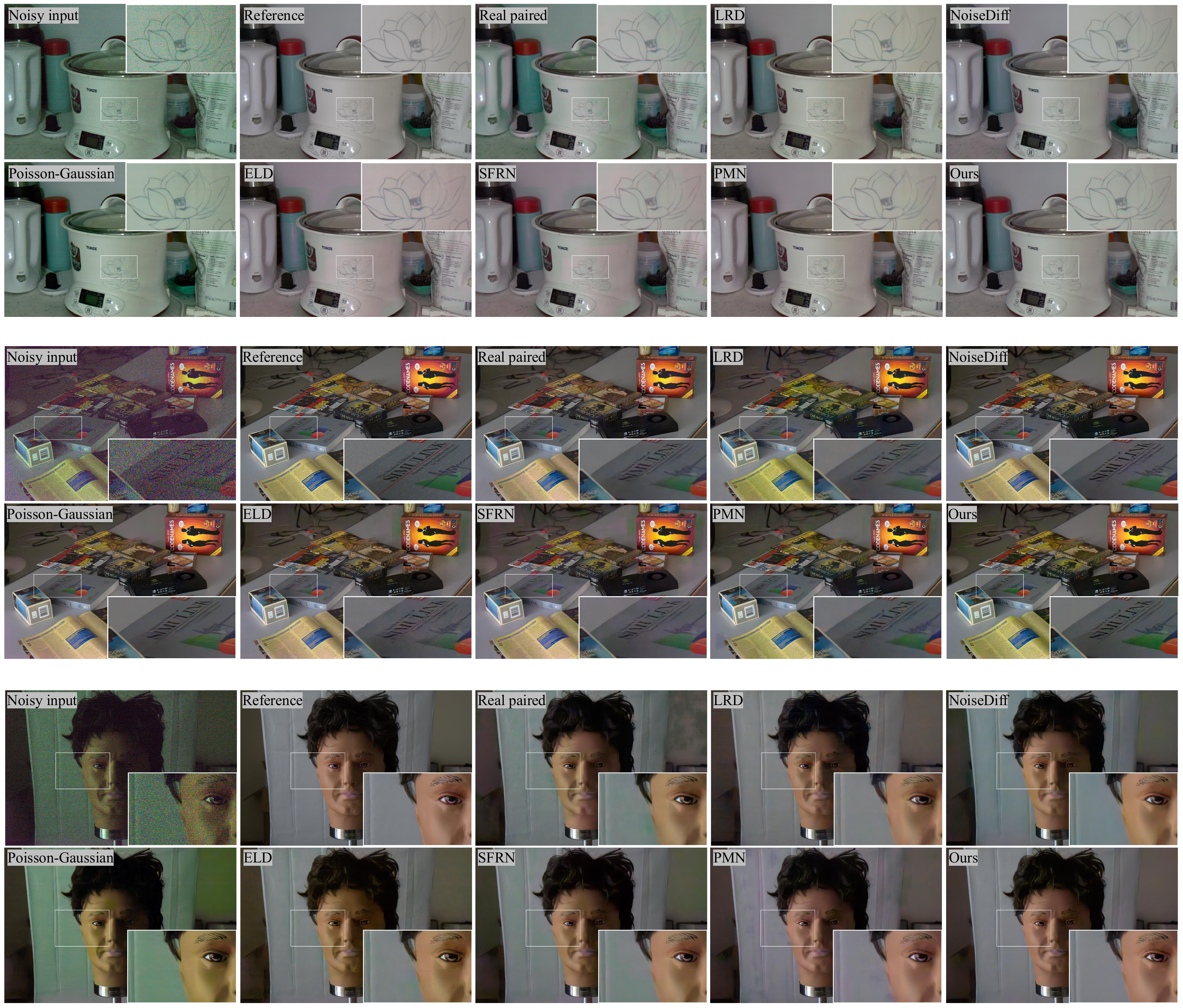}
\caption{Result comparison of denoisers trained on data synthesized using different methods on the SID test set.}
\label{fig:sid_1}
\end{figure*}

\begin{figure*}[t]
\centering
\includegraphics[width=\linewidth]{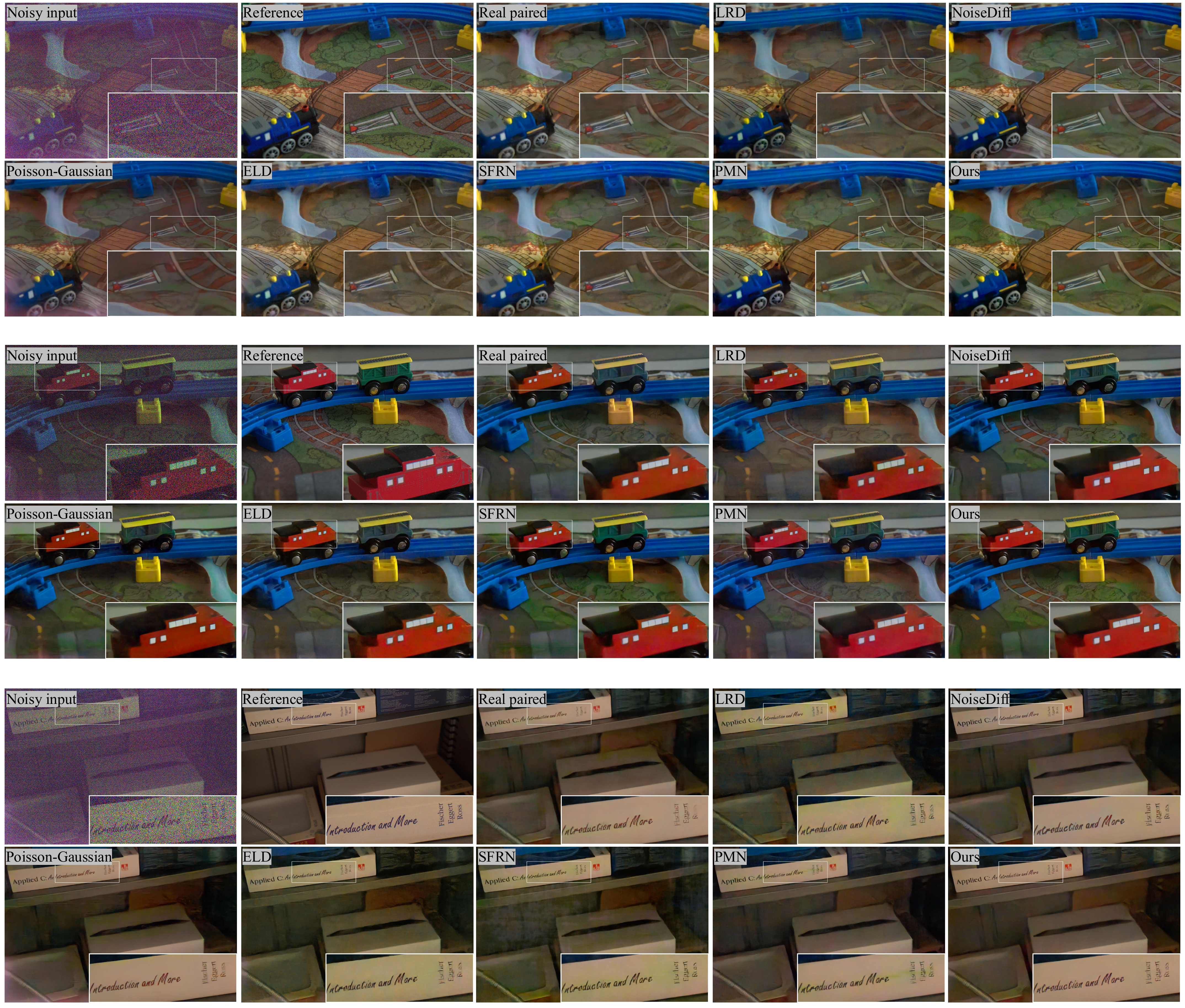}
\caption{Result comparison of denoisers trained on data synthesized using different methods on the SID test set.}
\label{fig:sid_2}
\end{figure*}

\begin{figure*}[t]
\centering
\includegraphics[width=\linewidth]{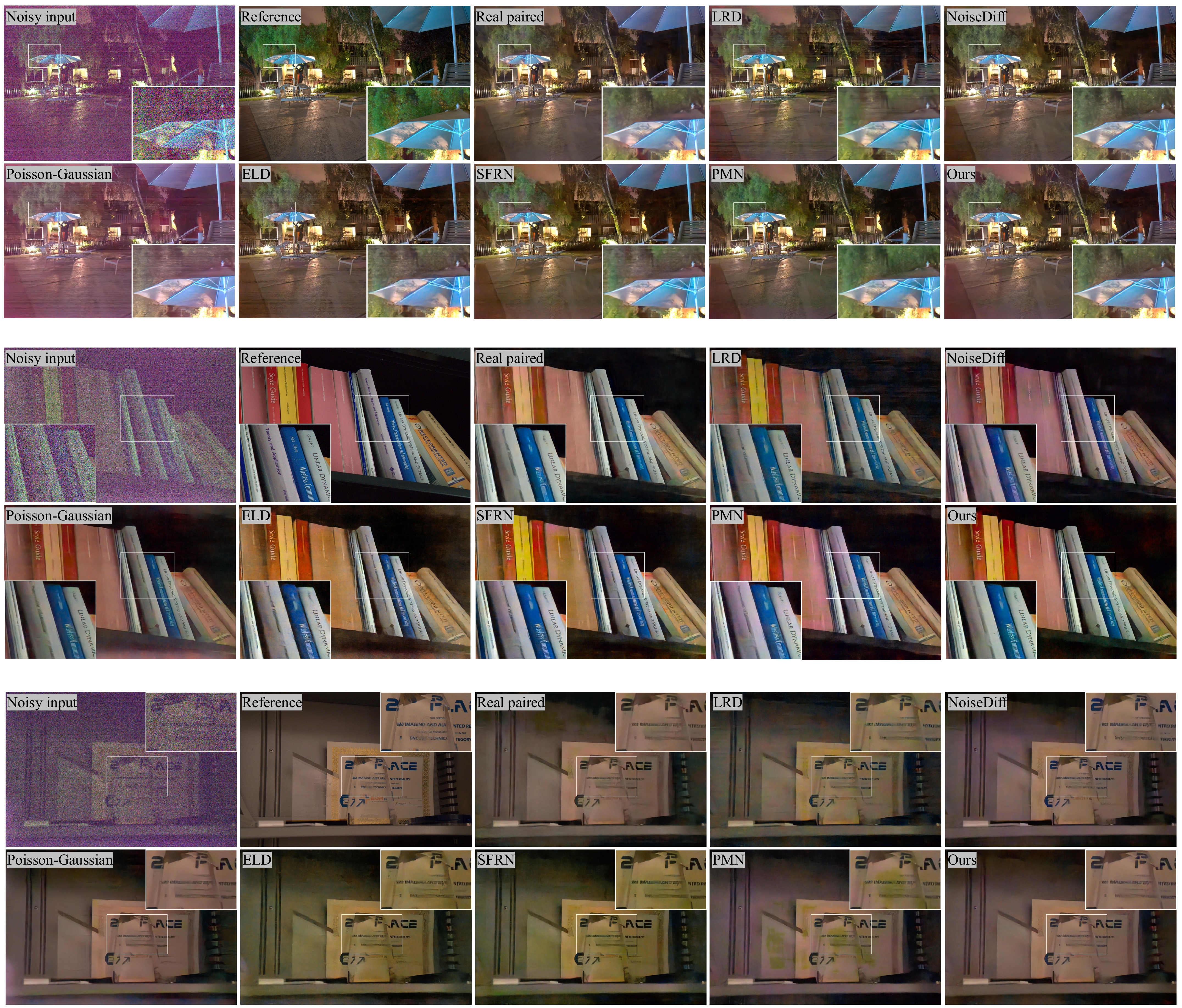}
\caption{Result comparison of denoisers trained on data synthesized using different methods on the SID test set.}
\label{fig:sid_3}
\end{figure*}

\begin{figure*}[t]
\centering
\includegraphics[width=\linewidth]{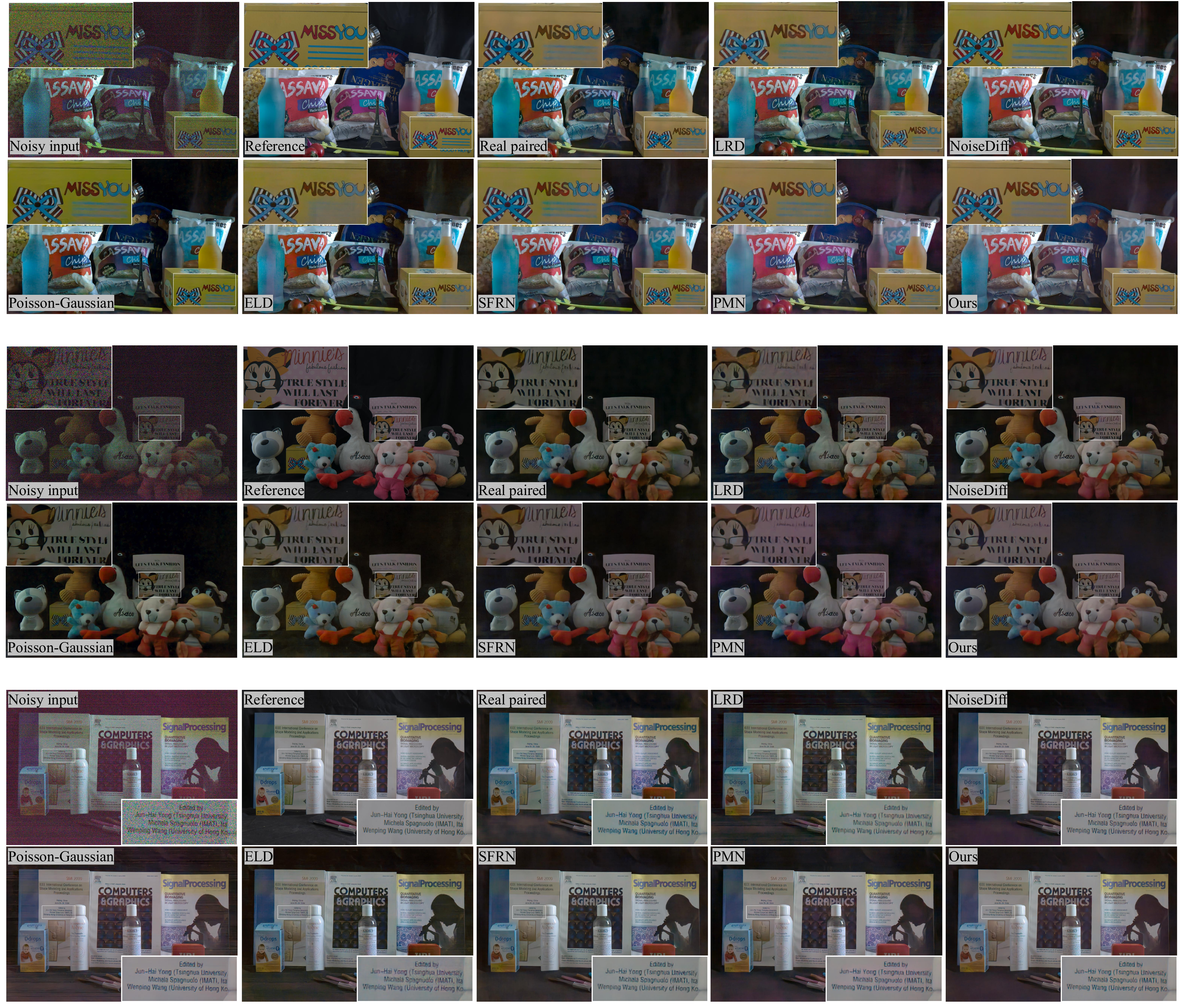}
\caption{Result comparison of denoisers trained on data synthesized using different methods on the ELD test set. Images are gamma-corrected ($\gamma=1.4$) for better visibility.}
\label{fig:eld_1}
\end{figure*}

\begin{figure*}[t]
\centering
\includegraphics[width=\linewidth]{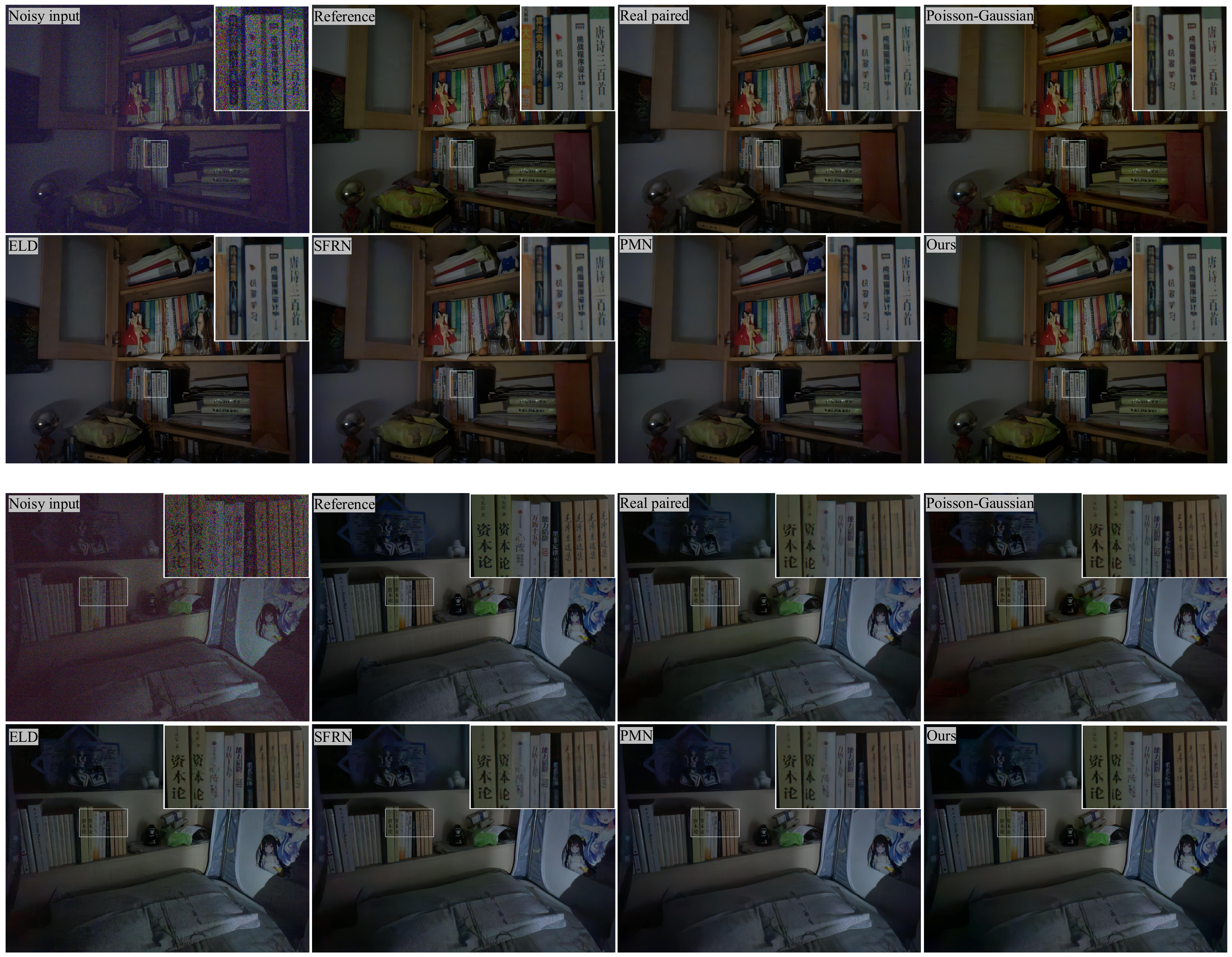}
\caption{Result comparison of denoisers trained on data synthesized using different methods on the LRID test set.}
\label{fig:lrid_1}
\end{figure*}

\begin{figure*}[t]
\centering
\includegraphics[width=\linewidth]{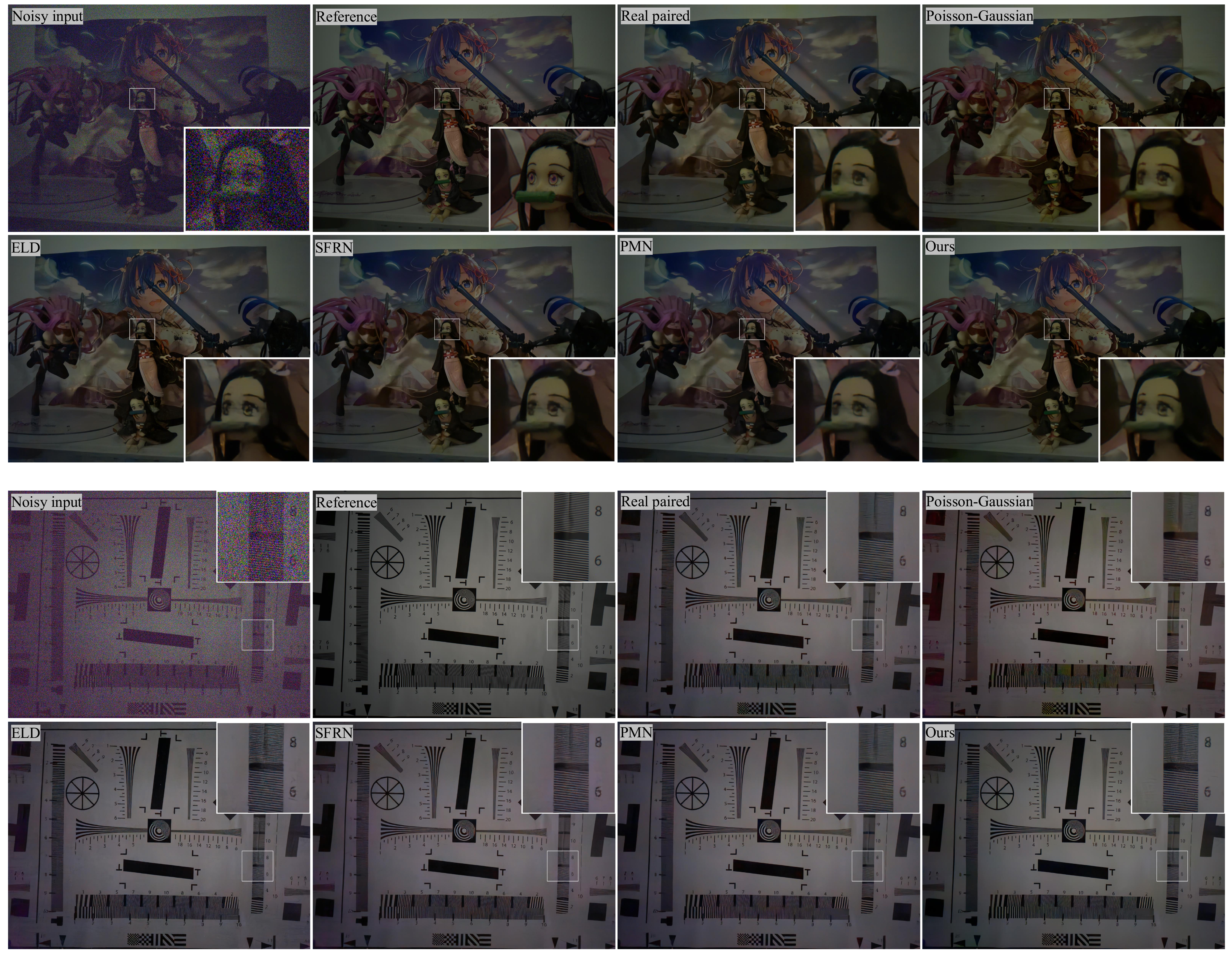}
\caption{Result comparison of denoisers trained on data synthesized using different methods on the LRID test set.}
\label{fig:lrid_2}
\end{figure*}

In this section, we provide more visual comparisons of denoising results from different denoisers trained on data synthesized using different methods. Fig.~\ref{fig:sid_1}, Fig.~\ref{fig:sid_2} and Fig.~\ref{fig:sid_3} are examples from the SID test set. Fig.~\ref{fig:eld_1} are examples from the ELD test set. 
Fig.~\ref{fig:lrid_1} and Fig.~\ref{fig:lrid_2} are examples from the LRID test set.

\end{document}